\def\year{2022}\relax
\documentclass[letterpaper]{article} 
\usepackage{aaai22}  
\usepackage{times}  
\usepackage{helvet}  
\usepackage{courier}  
\usepackage[hyphens]{url}  
\usepackage{graphicx} 
\usepackage{natbib}  
\usepackage{caption} 
\DeclareCaptionStyle{ruled}{labelfont=normalfont,labelsep=colon,strut=off} 
\usepackage{algorithm}
\usepackage{algorithmic}
\usepackage{amsmath}
\usepackage{amsfonts}
\usepackage{booktabs}
\usepackage{bm} 
\usepackage{subcaption}
\usepackage{natbib}
\usepackage{cite}
\usepackage{appendix}
\usepackage{newfloat}
\usepackage{listings}
\floatstyle{ruled}
\newfloat{listing}{tb}{lst}{}
\floatname{listing}{Listing}
\title{Concentration Network for Reinforcement Learning \\ of Large-Scale Multi-Agent Systems}

\author {
    Qingxu Fu\textsuperscript{\rm 1,2},
    Tenghai Qiu\textsuperscript{\rm 1,*}, 
    Jianqiang Yi\textsuperscript{\rm 1,2}, 
    Zhiqiang Pu\textsuperscript{\rm 1,2}, 
    Shiguang Wu \textsuperscript{\rm 1,2}
}
\affiliations {
    \textsuperscript{\rm 1} Institute of Automation, Chinese Academy of Sciences, Beijing 100190, China.\\
    \textsuperscript{\rm 2} School of Artificial Intelligence, University of Chinese Academy of Sciences, Beijing 100049, China.\\
    fuqingxu2019@ia.ac.cn, tenghai.qiu@ia.ac.cn, jianqiang.yi@ia.ac.cn, \\
    zhiqiang.pu@ia.ac.cn, shiguang.wu@outlook.com
}

\usepackage{bibentry}

\begin{document}

\maketitle

\begin{abstract}

When dealing with a series of imminent issues,
humans can naturally concentrate on a subset of these concerning issues by 
prioritizing them according to their contributions to motivational indices, 
e.g., the probability of winning a game.
This idea of concentration offers insights into reinforcement learning
of sophisticated Large-scale Multi-Agent Systems (LMAS) participated by hundreds of agents.
In such an LMAS, each agent receives a long series of entity observations at each step, 
which can overwhelm existing aggregation networks such as graph attention networks and cause inefficiency. 
In this paper, we propose a concentration network called ConcNet.
First, ConcNet scores the observed entities considering several motivational indices, 
e.g., expected survival time and state value of the agents,
and then ranks, prunes, and aggregates the encodings of observed entities 
to extract features.
Second, distinct from the well-known attention mechanism, 
ConcNet has a unique motivational subnetwork to explicitly
consider the motivational indices when scoring the observed entities.
Furthermore, we present a concentration policy gradient architecture 
that can learn effective policies in LMAS from scratch. 
Extensive experiments demonstrate that the presented architecture has excellent scalability and flexibility, 
and significantly outperforms existing methods on LMAS benchmarks.

\end{abstract}
\section{Introduction}

Multi-agent systems (MAS) from different areas have great disparity with each other,
considering the number of participating agents.
On one hand, MAS tasks like DouDizhu and Hide-and-Seek involve only
2 to 4 agents \cite{zha2021douzero, baker2019emergent}.
On the other hand,
other multi-agent challenges such as StarCraft Multi-Agent Challenge (SMAC),
Predators-Pray, Navigating, Attacker-Defender, and Formation-Control are usually
participated by 10 to 30 agents 
\cite{vinyals2017starcraft, lowe2017multi, chen2019crowd, deka2021natural, agarwal2019learning}.
In reinforcement learning (RL), 
these two types of problems are investigated from different perspectives with distinct methods.
Because the growth of the number of agents gradually shifts the core of the challenges
from agent-environment relationships to agent-wise interactions.
For simplicity, we refer to the former as mini-MAS, the latter as small-MAS.

\subsubsection{Large-Scale MAS.}
In this paper, we focus on the third type of MAS, 
which is participated by more than 100 agents.
The complexity of such large-scale MAS (LMAS)
increases quadratically as the agent number increases
considering the agent-wise relationships.
While LMAS creates the possibility for the emergence of very sophisticated 
cooperative behavior,
RL algorithms need to search a much large joint-policy space to 
find a satisfying policy.
So far, there has been little research on the RL algorithms of LMAS.
The related studies focus on oversimplified tasks such as 
gridworld dot-fighting and geometrical pattern-forming \cite{zheng2018magent,diallo2020multi, rubenstein2014programmable}.

An RL algorithm in LMAS faces following challenges:
\begin{itemize}
    \item \textbf{(C1) Partially observable environment.}
    An agent is influenced by a large number of entities, 
    which means the collection of other agents and non-agent elements in the environment.
    However, not all the state of entities are observable to this agent \cite{oliehoek2016concise}.
    \item \textbf{(C2) Decentralized execution.}
    The algorithm is restricted by the paradigm of centralized training with decentralized execution (CTDE),
    which is widely accepted since \cite{oliehoek2008optimal,kraemer2016multi}.
    \item \textbf{(C3) Huge observation space and observation uncertainty.}
    Each agent observes a long series of entities at each time step
    due to an exploding agent number. 
    Moreover, because of \textbf{(C1)}, the number of observable entities changes dynamically over an episode.
    Till now, to deal with this challenge,
    it is inevitable to use sequence modeling tools from the natural language processing (NLP) domain \cite{vaswani2017attention},
    such as attention mechanism and graph networks.
    \item \textbf{(C4) Scalability and dynamic agent number.}
    The algorithm should at least provide two kinds of scalability, namely, 
    the ability to adapt to scenarios that are diverse in initial agent number 
    before training (Training-Scalability) and after training (Testing-Scalability).
    Furthermore,  the available agent number changes dynamically during an episode 
    because most agents can be disabled or eliminated under certain conditions in LMAS.

\end{itemize}

\subsubsection{Bottlenecks of existing methods.}
Existing methods depend heavily on soft attention \cite{vaswani2017attention} 
to aggregate raw observations \cite{hoshen2017vain, iqbal2019actor}.
Graph attention networks, also relying on soft attention, 
can provide better performance in small-MAS tasks 
\cite{agarwal2019learning, jiang2018graph, deka2021natural}.
However, models established on soft attention suffer great performance degradation in LMAS tasks
because of the large agent number, the influence of \textbf{C3} as well as limitations of soft attention.

Soft attention uses score-softmax and weighted-sum procedures to extract features from a sequence of elements (meaning entity encodings here).
It allows encodings from essential entities to dominate the attentional output while keeping the network fully differential,
which is the key to its success in many applications including small-MAS.
Nevertheless, 
weights produced by the softmax function follows long-tail distributions after training \cite{zhou2021informer},
and hence small but non-zero weights are assigned to trivial entities unworthy of attention \cite{shen2018reinforced}.
In the case of LMAS, firstly,
large quantities of trivial entities weaken the attention given to the few truly essential entities.
Secondly,
agents in LMAS often need to consider multiple essential entities of a similar level of attention at the same time.
Unfortunately, the softmax function always magnifies the difference of attentional weights.
Even if the preceding network does produce identical attention weights,
the weighted-sum procedure will then degenerate into a naive average of feature vectors,
which erases the unique features of essential individual entities.
As a result, 
reinforcement learning of LMAS requires an alternative for soft attention mechanism.

Hard attention is such an alternative outside the boundary of RL but with valuable reference significance.
A similar problem that involves long-sequence attention calculation is long-sentence processing,
which is still a major challenge even in the well-explored NLP domain \cite{neishi2019relation}.
Studies have investigated hard attention for NLP and image caption \cite{xu2015show, shen2018reinforced}.
However, hard attention has a non-differentiable sample operation and is usually trained by REINFORCE \cite{williams1992simple}.
It creates a gap inside the policy network. In RL tasks, we lack additional rewards to fill this gap.

\subsubsection{Our contributions.}
We start by introducing a cognitive process known as concentration, 
or attention control \cite{astle2009using} in psychology, 
into the LMAS problems to meet the challenges.
Studies in psychology have in-depth discussions about the \textit{drive} of concentration (e.g., stimulus-driven or goal-driven).
But from our experience as humans, 
concentration can be simply considered as an ability to choose among a series of issues
and decide what to focus on or ignore according to one's \textit{motivation} or purpose.
Here we especially stress that the motivation plays an essential role in this process. 
For instance, 
we will prioritize sport-related issues over work-related issues if our primary motivation is winning a sport event,
or vice versa if our top motivation is catching deadlines at work.

We then model the process of concentration with neural networks.
Firstly, the representations of observed entities are evaluated by a score function 
according to several motivational indices, e.g., agent state value or expected survival time.
Secondly, the concentration network aggregates representations by ranking, pruning, concatenating and downsampling
to escape the aforementioned limitations of soft attention.
Thirdly, we solve the parameter differential problem with a unique motivation subnetwork,
which considers the motivational indices and supervises the training of the score function parameters.

Furthermore, we present a concentration policy gradient architecture designed based on the concentration network,
and demonstrate several possible architecture variants not only to show the flexibility of the concentration network,
but also to adapt LMAS with specific characters. 
E.g., entities have friend-or-foe distinction in competitive tasks, which is not the case in cooperative tasks.
We put forward an LMAS benchmark environment called Decentralised Collective Assault (DCA),
which simulates two-team competition combat participated by hundreds of agents.
We demonstrate the superior performance of the concentration-based architectures compared to existing alternative methods.
Ablation studies and further analyses are provided for a better understanding of our concentration network,
and to show the two types of scalability, namely the Training-Scalability and the Testing-Scalability.
\footnote{Conference paper at https://www.aaai.org/AAAI22Papers/AAAI-8368.FuQ.pdf}

\begin{figure*}[ht]
    \centering
    \includegraphics[width=.9\linewidth]{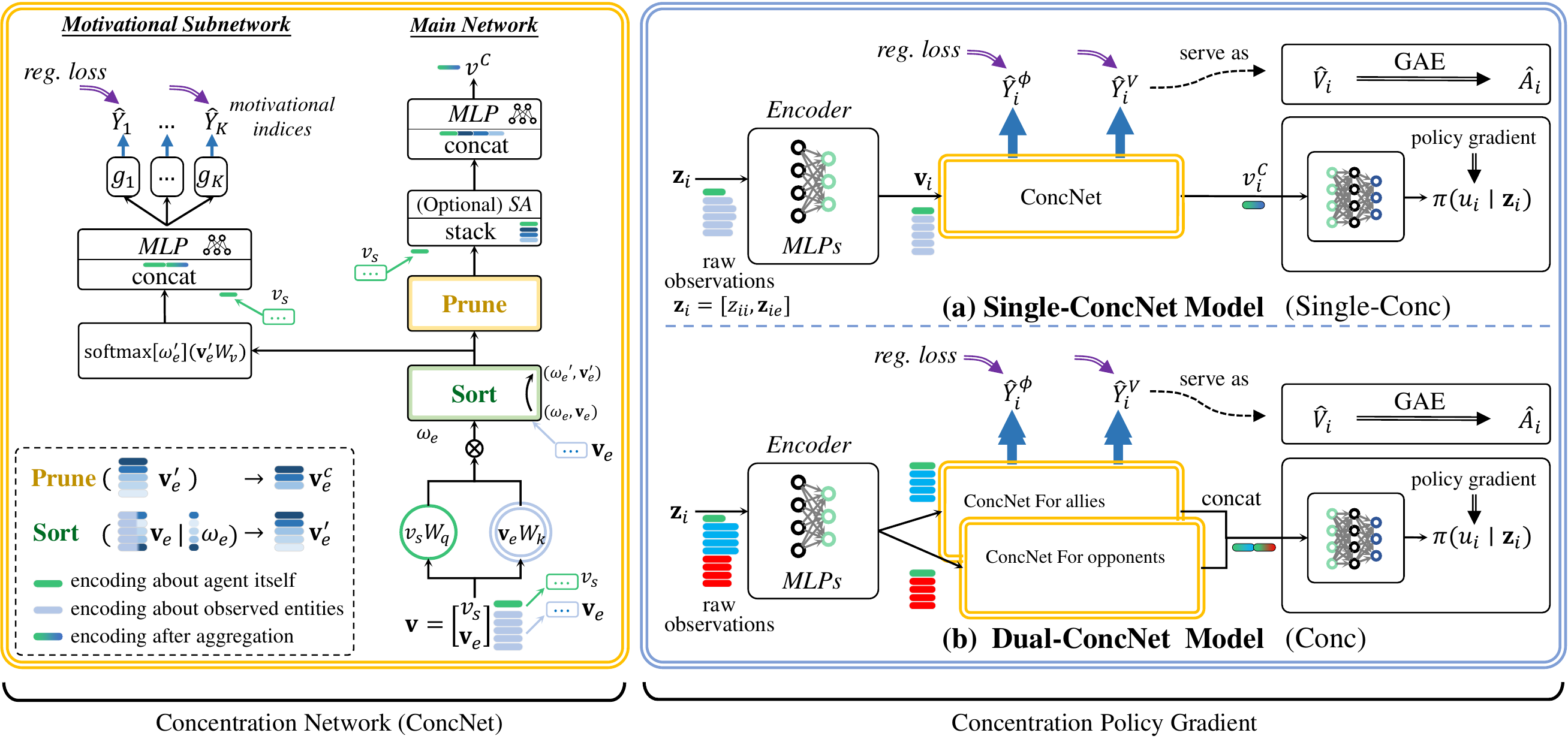}
    \caption{The structure of Concentration network and Concentration policy gradient.}
    \label{new-cct-fig}
\end{figure*}

\section{Entity-oriented Dec-POMDP}
An MAS task can be described as a Dec-POMDP \cite{oliehoek2016concise}.
However, there are two issues when applying Dec-POMDP in LMAS.
Firstly, a Dec-POMDP assumes a deterministic observation function,
but some LMAS tasks involve observation interference and noise that a Dec-POMDP cannot model.
Secondly, 
a Dec-POMDP integrates the self-observation and entity-observation of each agent into an integrated encoding,
which has high dimensions and uncertain length in LMAS due to the challenge \textbf{(C3)}.

We address these issues with an Entity-oriented Dec-POMDP (ED-POMDP).
The proposed ED-POMDP formulates LMAS with
a tuple $G=\langle A, E, \mathcal{U}, \mathcal{S}, P_t, \mathcal{Z}, P_o, r , \gamma\rangle$,
where $A=\lbrace 1,\dots,N \rbrace$ is a set of agents,
$E=\lbrace 1,\dots,N,\dots,M \rbrace$ is a set of entities, 
and $A \subseteq E$.
At each time step,
each agent $i \in \mathcal{A}$ chooses action $u_i \in \mathcal{U}$.
The joint action is represented as $\mathbf{u} \in \mathcal{U}^N$.
The true state of the environment is denoted as $s \in \mathcal{S}$.
The state transition function is $P_t(s'| s,\mathbf{u} ):\mathcal{S}\times\mathcal{U}\times\mathcal{S}\rightarrow [0,1]$.
An agent $i$ observes an entity $j\in E$ from observation $z_{ij} \in \mathcal{Z}$,
and the stochastic observation function is represented by $P_o(z,i,j| s):\mathcal{S}\times A \times E \times \mathcal{Z}$.
The chance of successfully observing the entity $j$ is $q_{ij}$,where
$q_{ij}(s)=\int_{z \in \mathcal{Z}} P_o(z,i,j| s) dz$.
We assume an agent can always observe itself, which means $q_{ii} =1$.
At each time step, 
the entities visible to agent $i$ are listed in $w(i | s)$.
The agent observations, denoted as $\mathbf{z}_i$, 
are represented by self-observation $z_{ii}$ and an entity-observation array $\mathbf{z}_{ie}$: 
\begin{equation}
\mathbf{z}_i=[z_{ii}, \mathbf{z}_{ie}], \operatorname{where}\; \mathbf{z}_{ie} = \left[z_{ij} \mid \forall j \in   w(i| \mathbf{s}),\;\operatorname{ s.t. } j \neq i \right]
\end{equation}
Finally, $r(i,s,\mathbf{u})\in \mathbb{R}$ is the reward function and  $\gamma$ is the discount factor.

Overall, the ED-POMDP has two major differences in comparison with Dec-POMDP. 
Firstly, the observation function is stochastic rather than deterministic. 
Secondly, an agent observes itself and each entity separately, instead of sealing them into a black-box encoding.

\section{Methods}

\subsection{Concentration Network}

We start by explicitly modeling a concentration process as follows:
\begin{enumerate}
    \begin{small}
    \item Given a set of entities $E$ and a set of motivational indices $M$.
    \item Rank entities in $E$ with a motivation-driven score function $R$. 
    \item Select top $d_c$ entities ($d_c \ge 1$) as subset $E' \subseteq E$.
    \item Aggregate the representations of entities in $E'$ by concatenation and downsampling.
    \end{small}
\end{enumerate}

E.g., for a pedestrian-avoidance driving policy, 
each pedestrian is denoted as $e \in E$.
The probability of collision makes one of the motivational indexes.
The motivation-driven score function $R$ selects $d_c$ pedestrians 
contributing most to the collision probability into $E'$,
determining the preference of concentration. 
Finally, the representations in $E'$ are aggregated for further processing.

Next, we realize this concentration process with neural networks
and refer to it as a Concentration Network (ConcNet).
As shown in the left part of Fig.~\ref{new-cct-fig},
a ConcNet has two parts, a straightforward \textbf{main network} and a special \textbf{motivational subnetwork}.

\subsubsection{Concentration Main Network.}
As a prerequisite, raw observations need to be encoded into a feature space with dimension $d_k$ before 
fed into ConcNet.
Let $d_e$ be the number of observed entities.
Encoded observations of entities are represented with matrix $\mathbf{v}_e \in \mathbb{R}^{d_e \times d_k}$.
The encoded self-observation is denoted as $v_s \in \mathbb{R}^{1 \times d_k}$.

First, the \textit{\textbf{score}} function $R$ uses a dot product to calculate scores $\omega_e$. 

\begin{equation}
    \omega_e = R(v_s, \mathbf{v}_e) = \frac{(\mathbf{v}_e W_k )(v_s W_q)^T }{\sqrt{d_k}}
    \label{eq:score}
    \end{equation}
where $W_q$ and $W_k\in \mathbb{R}^{d_k \times d_k}$ are learnable parameter matrices.
Next,
the observed entity representations, namely $\mathbf{v}_e$, are \textit{\textbf{ranked}}
based on their scores $\omega_e \in \mathbb{R}^{d_e}$ by switching matrix rows
(rows with top scores are placed at the top):
\begin{equation}
    \begin{array}{l}
            \omega_e'= \operatorname{Sort} \left(\omega_e \right), \;
            \mathbf{v}_e' = \operatorname{Sort}\left(\mathbf{v}_e|\omega_e \right)
   \end{array}
\end{equation}
where the sorted versions of $\mathbf{v}_e$ and $\omega_e$ are represented as $\mathbf{v}_e'$ and $\omega_e'$.

Afterward, in a way that resembles hard attention \cite{xu2015show},
we \textit{\textbf{prune}} $\mathbf{v}_e'$ by
selecting the top $d_c$ rows and removing the rest $(d_e-d_c)$ rows, 
resulting in $\mathbf{v}_e^c$:
\begin{equation}
    \begin{array}{l}
        \mathbf{v}^c_e = \operatorname{Prune}\left( \mathbf{v}_e' | d_c  \right),\ \mathbf{v}^c_e \in \mathbb{R}^{d_c \times d_k}
   \end{array}
\end{equation}
In case of $d_e < d_c$, we deal with it using zero-paddings.

For the final aggregation procedure,
ConcNet concatenates $v_s$ with flattened $\mathbf{v}^c_e$ into $v^\ell \in \mathbb{R}^{(d_c+1) \times d_k}  $.
Then an MLP layer $f_{cn}(\cdot)$ is used to restore the dimension of representation back to $d_k$ (downsample):
\begin{equation}
    \begin{array}{rl}
        v^c &= f_{cn}\left( v^\ell  \right), \ v^c \in \mathbb{R}^{d_k} 
    \end{array}
\end{equation}
where $v^c$ is the output of ConcNet main network.
Additionally, 
we give another network variant by inserting an optional self-attention (SA) layer right before the concatenation step.

\subsubsection{Motivational Subnetwork.}

ConcNet is incomplete without the guidance of motivation indices $M$.
Note that the parameters of the score function $R$ are not yet differential due to the rank-and-prune operation.
We address this problem by designing a Motivational Subnetwork as shown in Fig.~\ref{new-cct-fig},
which is designed to be differential w.r.t. the parameters of $R$.

This subnetwork will be used to predict each motivational index $m\in M$,
in which process the regression loss is back-propagated to supervise the training of $R$,
making the score function sensitive to entities that are concerned with $M$.
In other words, the score function is trained to recognize entities 
with the most significant impact on the motivational indices in this subnetwork, 
and to correct the concentration preference of ConcNet.

The motivational indices $M$ are no doubt essential in this subnetwork.
E.g., agent state value can be used as a motivational index, because the core motive in RL is the reward.
Also, the expected time of survival can make another motivational index, 
since an agent has to be alive to do anything at all.
Multiple indices can co-exist, and the number of chosen motivational indices
is denoted as $K$.

An overview of the design of the motivational subnetwork is shown in Fig.~\ref{new-cct-fig}.
The subnetwork begins at its divergence from the main network, 
right after the ranking operation.
Since the output of this subnetwork has no forward influence on the main network,
it is safe to use softmax and weighted-sum for feature extraction here:
\begin{equation}
    \label{eq:softmax}
    v^m_e = \operatorname{softmax}\left[ \omega_e' \right] \cdot (\mathbf{v}_e' W_v)
\end{equation}
where $W_v$ is another learnable parameter matrix.

Then the result $v^m_e$ of Eq.~\ref{eq:softmax} is concatenated with $v_s$ by a skip connection,
before an MLP layer $f_{m}(\cdot)$ concludes a motivational representation $v^m$:
\begin{equation}
    \label{eq:motivation-cat}
    v^m = f_{m}\left[\operatorname{concat}\left( v_s, v^m_e  \right) \right]
\end{equation}

Next, the subnetwork estimates $K$ motivational indices with MLP networks $\lbrace g_1, \dots, g_K \rbrace$:
\begin{equation}
    \hat{Y}_k = g_k(v^m ),\; k \in \lbrace 1,\dots, K \rbrace
    \label{eq:motivation-mlp}
\end{equation}
The estimated indices are denoted as $\lbrace \hat{Y}_1, \dots, \hat{Y}_K \rbrace$, 
and the regression loss function is summarized by $\mathcal{L}_{reg}$:
\begin{equation}
\mathcal{L}_{reg}^{\theta_r} =  \mathcal{L}_{reg}^1 + \dots + \mathcal{L}_{reg}^K = \sum^{K}_{k=1} \mu_k \mathbb{E}_{\mathcal{D}} \left[ \left\| \hat{Y}_k - Y_k \right\|^{2} \right]
\label{eq:Loss-reg}
\end{equation}
where $Y_k$ is the true value of the $k$-th motivational index, 
$\mathcal{D}$ is a collection of most recent episode samples,
and $\lbrace \mu_1,\dots, \mu_k \rbrace$ are constant scalars that weigh the importance of each motivation.
The parameter collection $\theta_r$ includes parameters of the motivational subnetwork and parameters of $R$.

\subsection{Concentration Policy Gradient}
Now we put forward a concentration policy gradient architecture
as an example application of ConcNet.
\subsubsection{Motivational Indices Selection}
The selection of motivational indices is important.
While there are many choices of indices when considering specific tasks,
two general indices exist in LMAS problems,
namely the state value $Y^{V}_i$ and the expected time of survival $Y^{\phi}_i$ 
mentioned in the previous section.
For generality and expansibility, 
the concentration policy gradient architecture is only established on these two motivation indices.

The regression loss from the former motivation index $Y_{1,i}=Y^{V}_i$ is $\mathcal{L}_{reg}^1$:
\begin{equation}
\begin{aligned}
    \mathcal{L}_{reg}^1 &= \mu_1 \mathbb{E}_{\mathcal{D}}
                            \left[ \left\| \hat{Y}^{V}_i(\mathbf{z}_{i}) - 
                            Y^{V}_i \right\|^{2}     \right] \\
    Y^{V}_i &= \mathbf{r}_{i} = \sum_{l=0}^{\infty} \gamma^{l} r_i(t+l)
    \label{eq.mc-motivation}
\end{aligned}
\end{equation}
where $\hat{Y}^{V}_i$ is the estimated index reflecting agent's state value,
$\mathbf{r}_{i}$ is the discounted sum of rewards.
Strictly speaking, Eq.~\ref{eq.mc-motivation} utilizes the Monte Carlo approach \cite{sutton1998introduction}
to train the state value estimation.

As for the latter motivation index $Y_{2,i}=Y^{\phi}_i$, 
we define it as a truncated survial-time countdown:
\begin{equation}
\begin{aligned}
    Y^{\phi}_i(t; T_{max}) 
    =         
    min \left[T_{i}-t,T_{max} \right], \; t \in [0, T_i]
\end{aligned}
\end{equation}
where $T_{i}$ is the total survival time of agent $i$ in current episode and 
$T_{max}$ is a threshold to limit $Y^{\phi}_i \in [0,T_{max}]$.
Correspondingly, 
a small adjustment is made to the loss function $\mathcal{L}_{reg}^2$ by
replacing $\mathcal{D}$ with $\mathcal{D}'$, 
where $\mathcal{D}'\subseteq \mathcal{D}$ contains samples satisfying $t\in [T_i - T_{max}, T_i]$.
The loss function $\mathcal{L}_{reg}^2$ is:
\begin{equation}
    \mathcal{L}_{reg}^2 = \mu_2 \mathbb{E}_{\mathcal{D}'}
                            \left[ \left\| 
                            \hat{Y}^{\phi}_i(\mathbf{z}_{i}) -  Y^{\phi}_i
                            \right\|^{2}     \right], 
\end{equation}

\subsubsection{Structure}
In the right part of Fig.~\ref{new-cct-fig}, 
we present concentration policy gradient models (a) and (b) established upon this 2-motivation ConcNet.
Model (a) is referred to as Single-ConcNet and (b) as Dual-ConcNet.


\newcommand{\diiv}{.45}
\newcommand{\dwhv}{1}
\newcommand{\mycenter}{\centering}
\begin{figure}[!t]
    \centering
    \begin{subfigure}[t]{\diiv\linewidth}
    \centering
    \includegraphics[width=\linewidth,height=\linewidth]{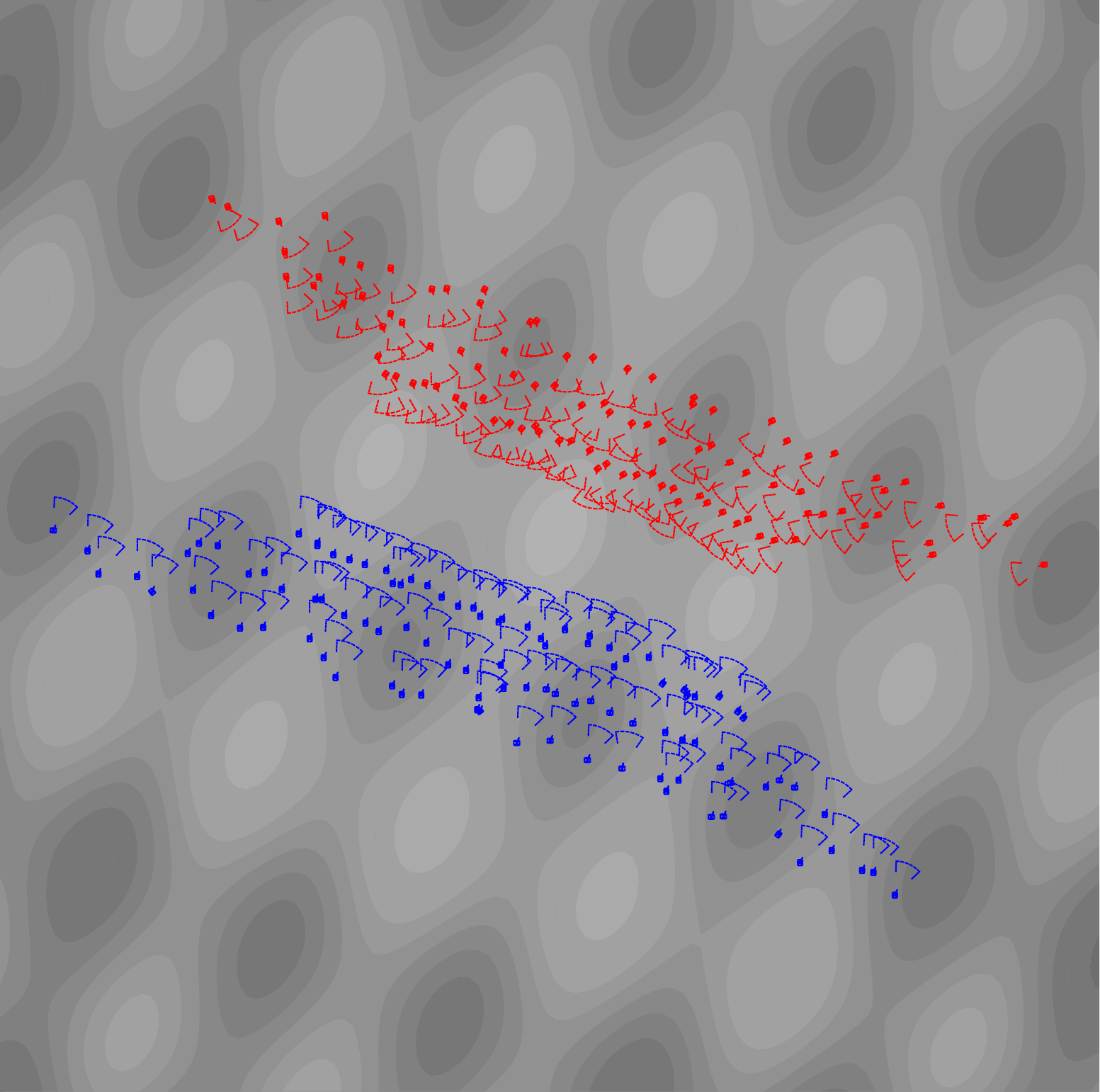}
    \caption{$N_{blue}=N_{red}=100$.}
    \label{fig:dca1}
  \end{subfigure}
  \begin{subfigure}[t]{\diiv\linewidth}
    \centering
    \includegraphics[width=\linewidth,height=\linewidth]{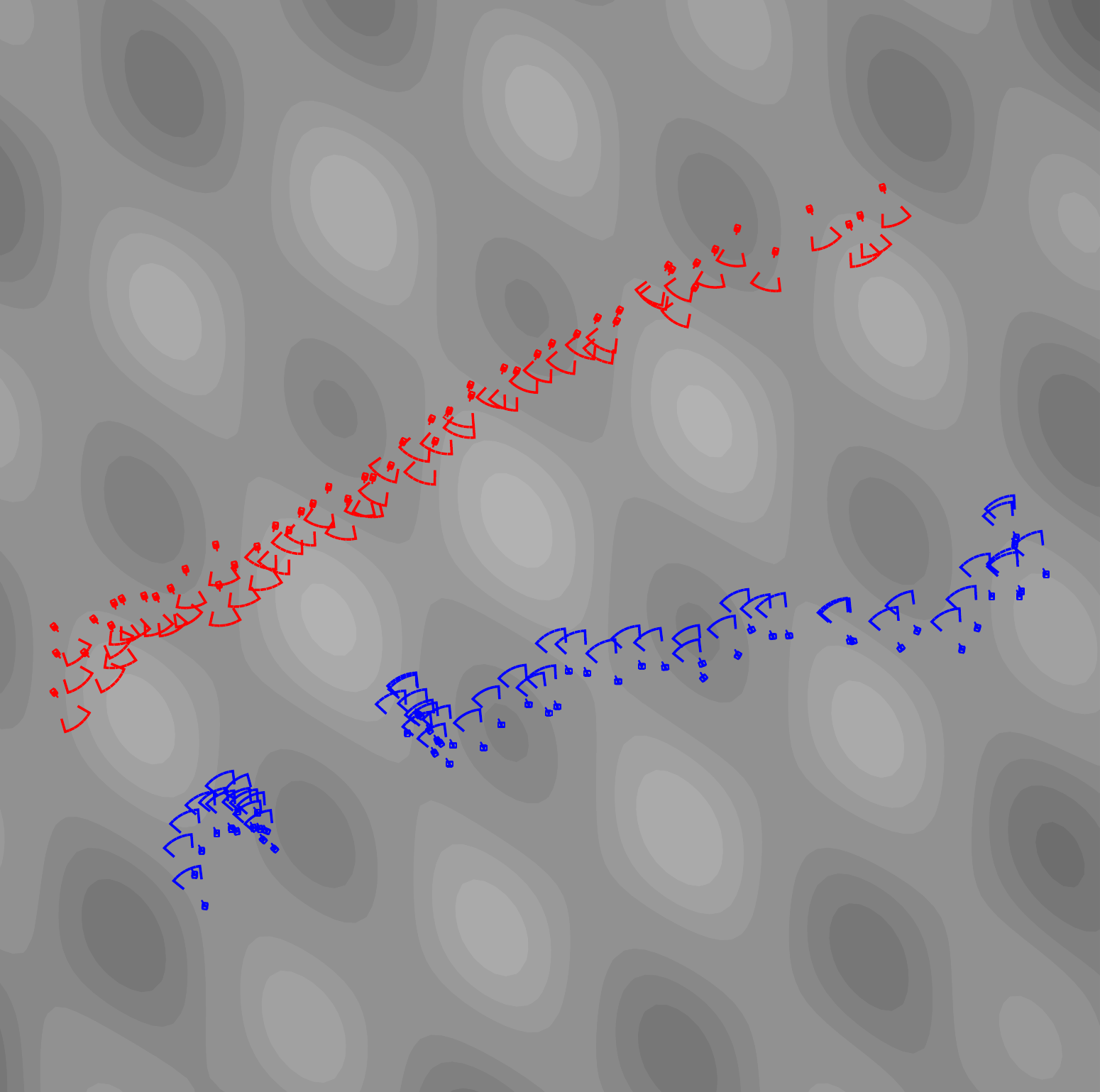}
    \caption{$N_{blue}=N_{red}=50$.}
    \label{fig:dca2}
  \end{subfigure}\\
  \begin{subfigure}[t]{\diiv\linewidth}
    \centering
    \includegraphics[width=\linewidth,height=\linewidth]{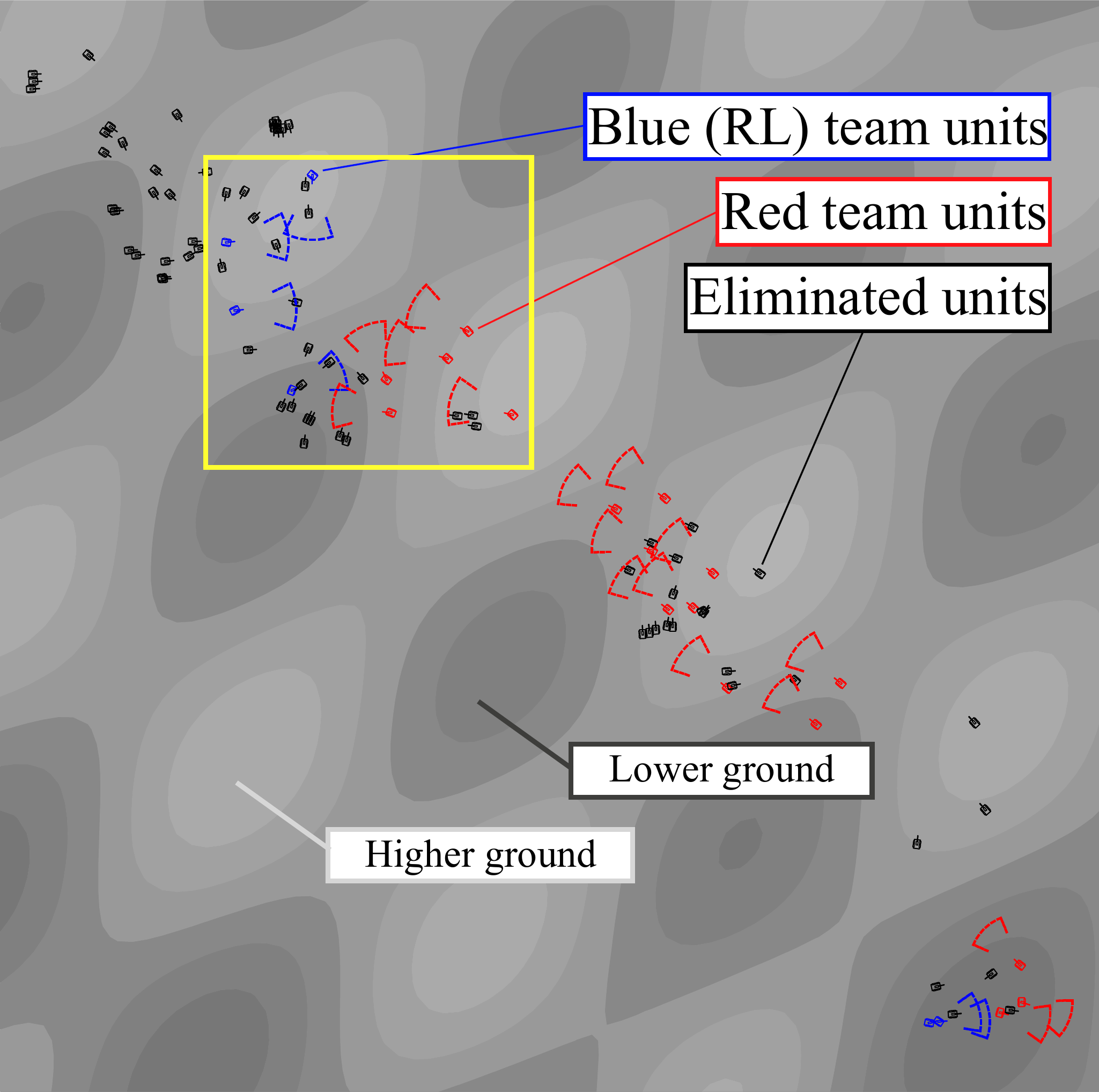}
    \caption{Middle of a game.}
    \label{fig:dca3}
  \end{subfigure}
  \begin{subfigure}[t]{\diiv\linewidth}
    \centering
    \includegraphics[width=\linewidth,height=\linewidth]{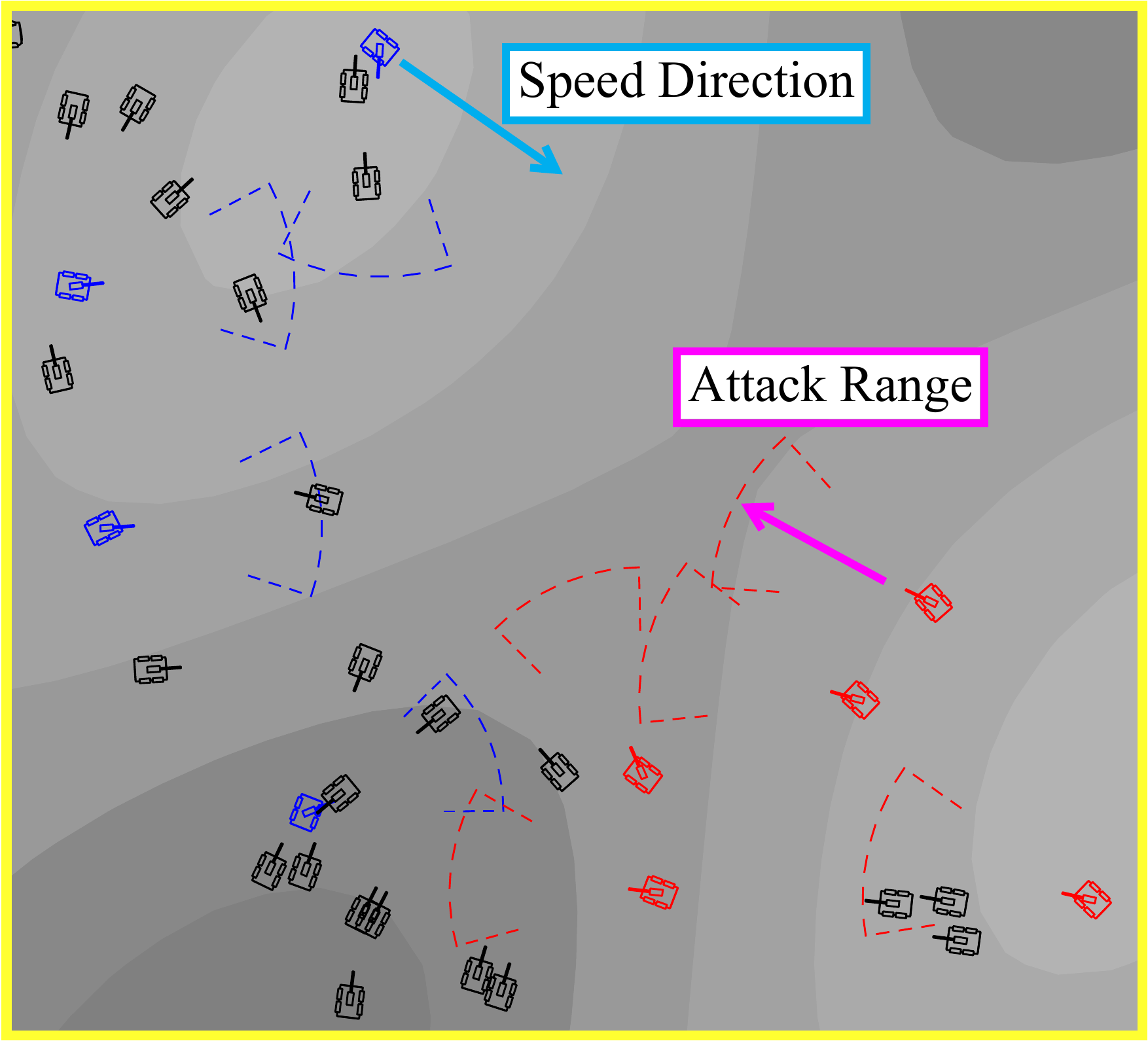}
    \caption{Zoom in (c).}
    \label{fig:dca4}
  \end{subfigure}
  \caption{The Decentralised Collective Assault (DCA) environment supports the cooperative
  competition between two teams of agents. The terrain is represented by gray contours.}
  \label{fig:dca}
\end{figure}
 
\begin{figure}[h]
  \centering
  \begin{subfigure}[t]{\linewidth}
    \centering
    \includegraphics[width=\linewidth]{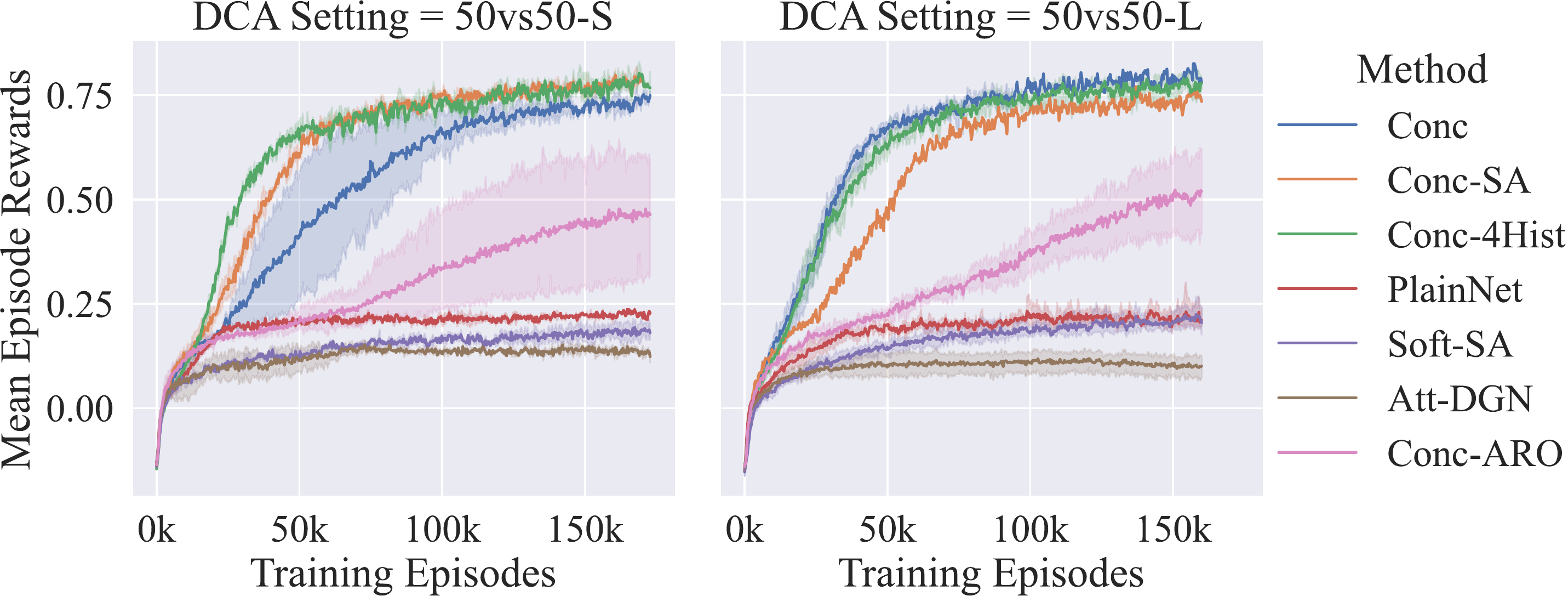}
    \caption{Mean Episode Reward.}
    \label{fig:comp1}
  \end{subfigure} \\
  \begin{subfigure}[t]{\linewidth}
    \centering
    \includegraphics[width=\linewidth]{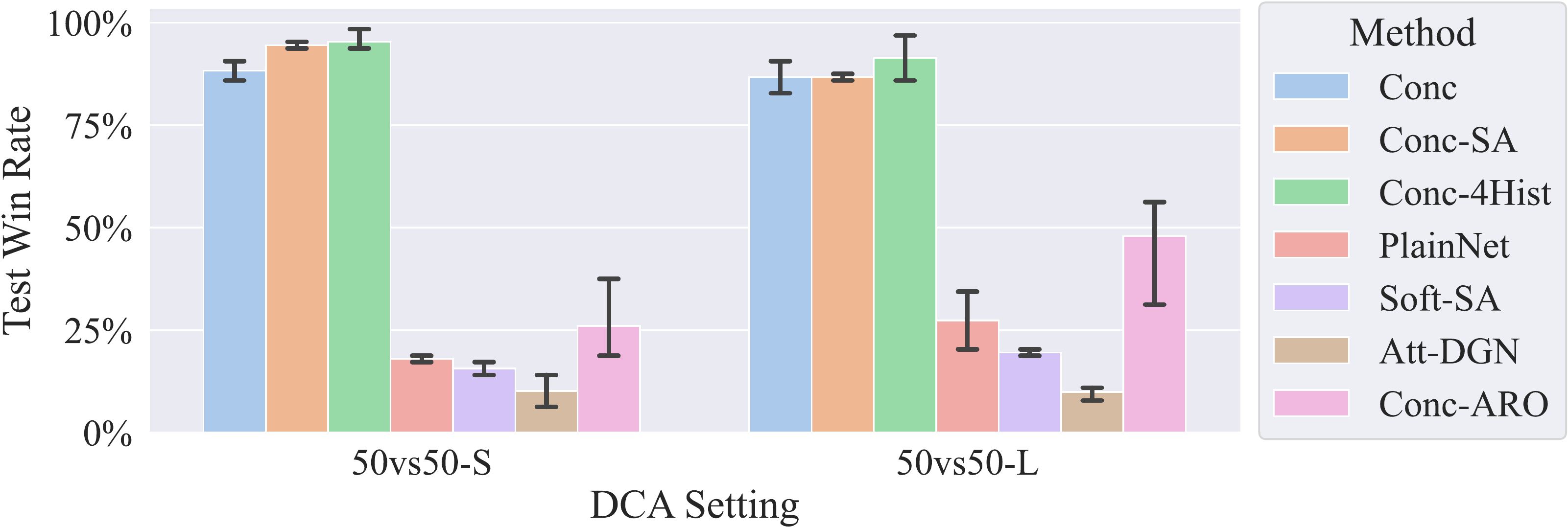}
    \caption{Win Rate.}
    \label{fig:comp2}
  \end{subfigure} 
  \caption{A comparison study between concentration-based model and attention-based model.}
  \label{fig:comp}
\end{figure}

Single-ConcNet is a basic and straightforward model to apply ConcNet.
It begins by encoding raw observations with MLPs, followed by a ConcNet.
The outputs of ConcNet's main network are used to estimate action distributions $\pi(u_i | \mathbf{z}_i)$.
Therefore, the main network is necessary for both the training stage and the testing stage.
In comparison,
the motivation subnetwork of ConcNet is only needed during training, 
playing an indispensable role in shaping the score function $R$.
This subnetwork is also responsible for 
providing motivational index $\hat{Y}^{V}_i$ as $\hat{V}_i$ for advantage estimation.
Once the training is done, the entire motivation subnetwork is no longer needed.

As an extension of Single-ConcNet, 
Dual-ConcNet model in Fig.~\ref{new-cct-fig}(b) uses two ConcNets to deal with problems concerning multiple types of entities.
For instance, 
in tasks with known \textit{friend-or-foe} entity identifications, 
a Dual-ConcNet model (referred to as Conc for simplicity) can process ally and 
opponent entities separately for better performance and flexibility.

Furthermore, 
when history observation is essential for making decisions,
a Dual-ConcNet model can also be used to process \textit{past-or-present} observations, 
serving as an alternative choice besides RNN.
More specifically, in this case,
we introduce a FIFO (first-in, first-out) memory pool to store history observations.
Then one of the ConcNet in (b) is used to process present observations,
with the other ConcNet to process past observations taken from the FIFO pool.
This variant considering history observations
is referred to as Conc-4Hist to distinguish from the Conc model shown in Fig.~\ref{new-cct-fig}(b).

As another difference with the Single-ConcNet model,
the two motivation subnetworks in Fig.~\ref{new-cct-fig}(b) are merged internally by the concatenation of Eq.~\ref{eq:motivation-cat} 
to produce joint motivational indices.
The implementation details of Conc and Conc-4Hist are discussed in the Appendix.
Note that one of the motivational indexes, 
namely the state value index $Y^{V}_i$, 
also participates in the estimation of advantage $\hat{A}_{i}$ to calculate the policy gradient.

Advantage $\hat{A}_{i}$ is estimated with Generalized Advantage Estimation (GAE) \cite{schulman2015high},
and $\lambda$ is the GAE parameter.
\begin{equation}
    \hat{A}_{i}(t)=
    \sum_{l=0}^{\infty}(\gamma \lambda)^{l} 
    \left[
        r_{i}(t)
        +\gamma \hat{V}_{i}\left(t+1\right)
        -\hat{V}_{i}\left(t\right) 
    \right]
\end{equation}
\begin{equation}
\label{eq:pg}
\nabla_{\theta_\pi} J\left(\pi\right)
=
\mathbb{E}_{\mathcal{D}}
\left[
      \nabla_{\theta_\pi} \log \pi\left(
          u_{i} | \mathbf{z}_{i}, \theta_\pi
        \right)\hat{A}_{i}
\right]
\end{equation}
\begin{equation}
\label{eq:reg-repeat}
\underset{\theta_{r}}{\operatorname{minimize}} \; \mathcal{L}_{reg}^{\theta_{r}}
\end{equation}
where $\theta_\pi$ is the collection of policy parameters.

Eventually, the advantage is used to calculate the policy gradient (Eq.~\ref{eq:pg}), 
which updates the policy.
In our model, by solving the regression problem in Eq.~\ref{eq:reg-repeat},
two goals are achieved at one stroke: 
firstly, an estimator of the value function is trained;
secondly, the score function $R$ is trained to consider motivational indices.

\section{Experiments}

\begin{figure}[h]
  \begin{subfigure}[t]{\linewidth}
    \centering
    \includegraphics[width=\linewidth]{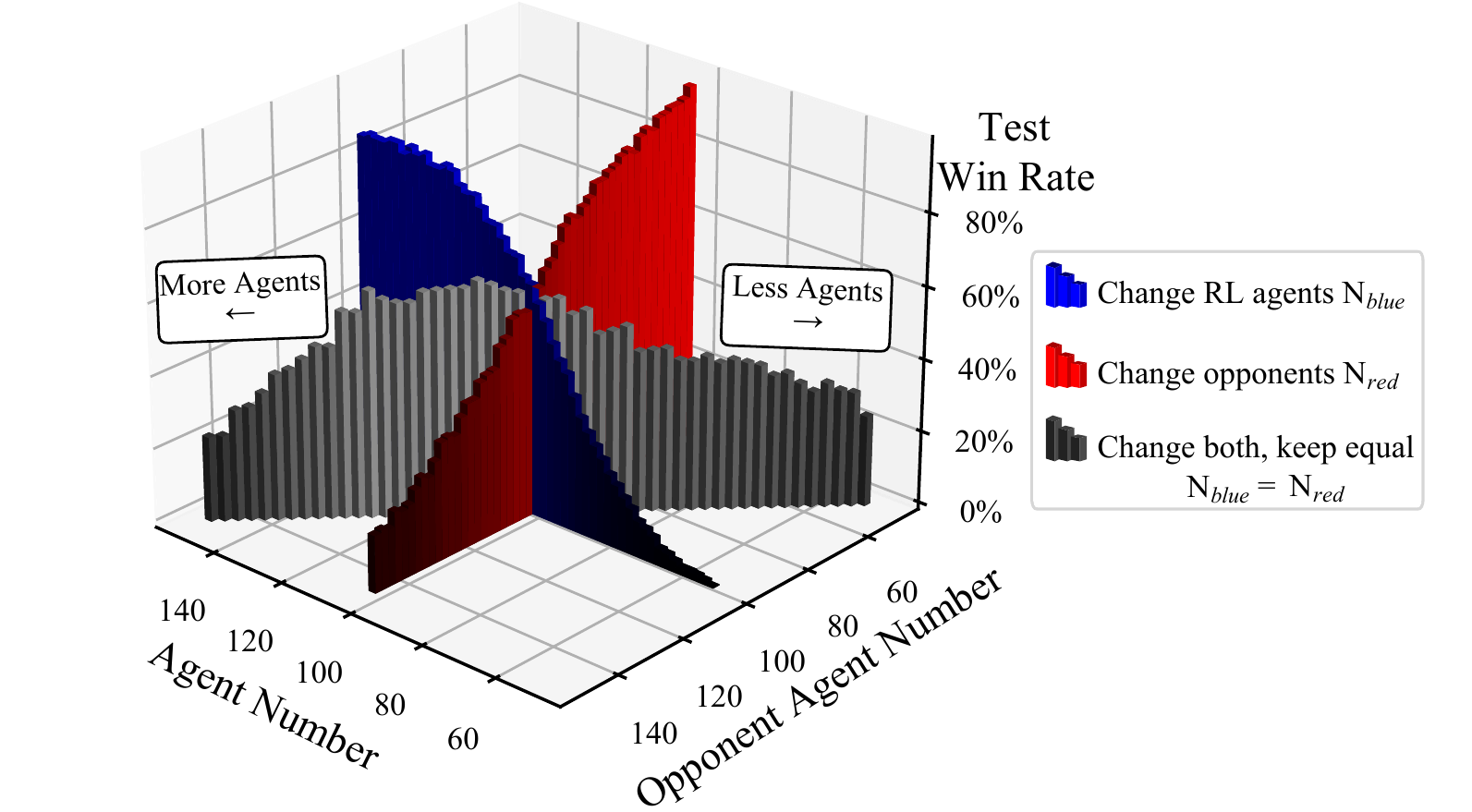}
    \label{fig:abl-scale1}
    \caption{Scalability of Conc Model.}
  \end{subfigure} 
  \begin{subfigure}[t]{\linewidth}
    \centering
    \includegraphics[width=.9\linewidth]{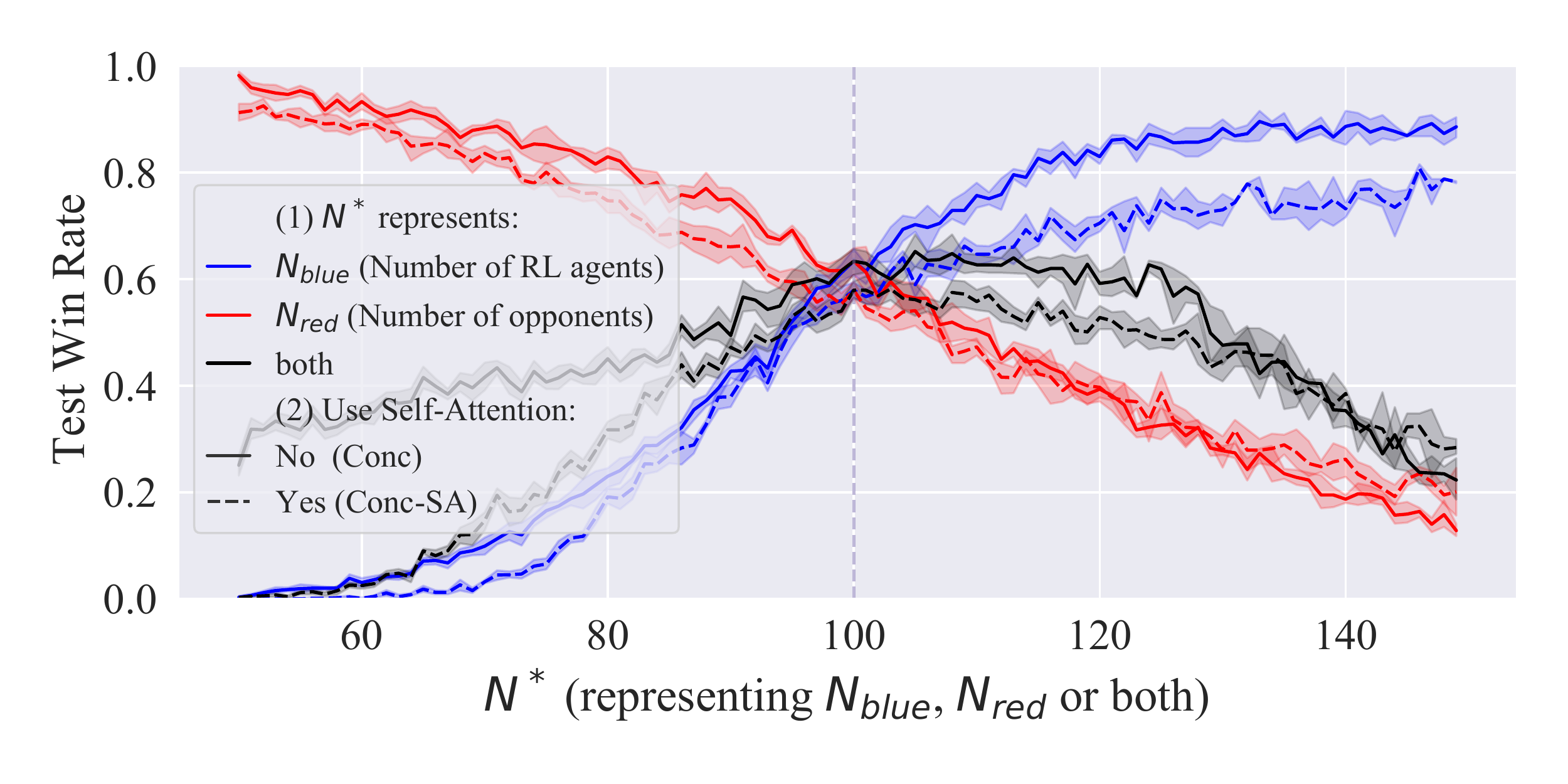}
    \label{fig:abl-scale2}
    \caption{Comparing the testing-scalability of Conc and Conc-SA.}
  \end{subfigure} 

  \caption{Testing-scalability of our models. 
  Conc and Conc-SA models are trained under $N_{blue}$=$N_{red}$=$100$,
  and then tested under a series of different settings:
  (1)
  $N_{blue}\in[50,150]$ with $N_{red}=100$.
  (2)
  $N_{red}\in[50,150]$ with $N_{blue}=100$.
  (3) 
  $N_{blue}=N_{red}\in[50,150]$.}
  \label{fig:abl-scale}
\end{figure}

\begin{figure}[!th]
  \centering
  \includegraphics[width=\linewidth]{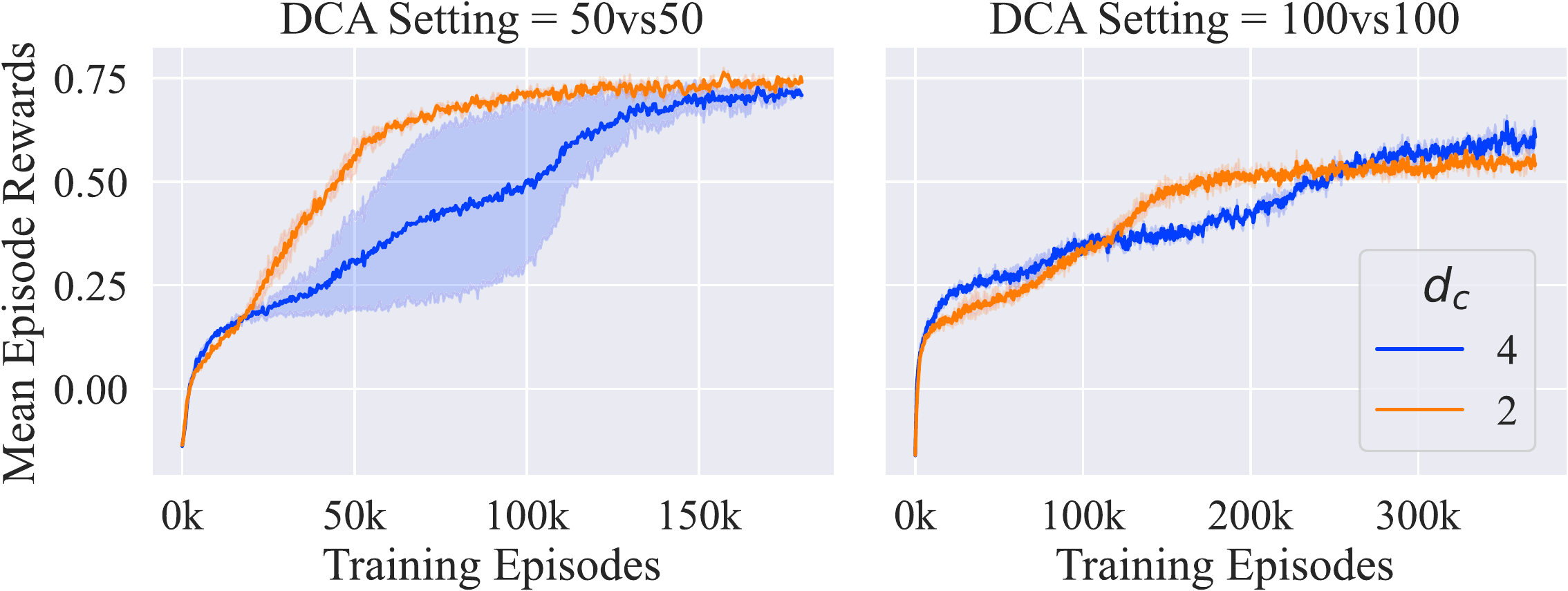}
  \caption{Performance comparison of our Conc model under settings of 100vs100 and 50vs50.}
  \label{100vs100}
\end{figure}

\begin{figure}[h]
  \centering
  \includegraphics[width=\linewidth]{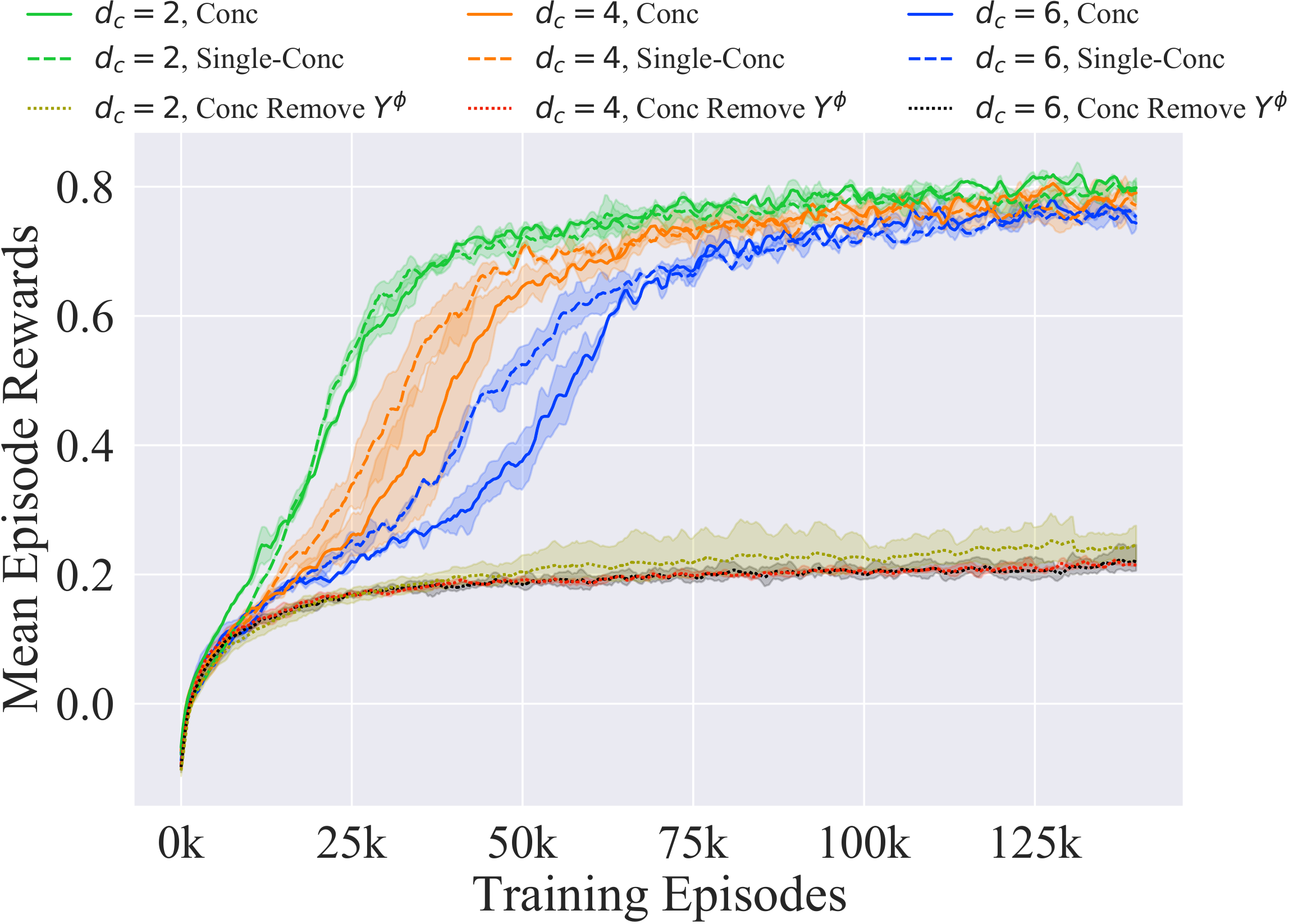}
  \caption{The influence of $d_c$ and the performance of Dual-ConcNet and Single-ConcNet in the 50vs50-S setting.
  By increasing $d_c$ or adopting the Dual-ConcNet model, more trainable network parameters are
  introduced and the training is slightly slowed down. 
  Neither factor has a significant impact on the final episode rewards.
  }
  \label{fig:abl-dc}
\end{figure}

In this section, we will illustrate the efficiency of the ConNet and the concentration policy gradient under complex LMAS tasks in comparative studies.
Ablation studies and further analyses are provided for a better understanding of our method, 
and to demonstrate the flexibility and scalability of ConNet.
\footnote{The source code is available at the following repository. https://github.com/binary-husky/hmp2g/tree/aaai-conc.
}

\subsection{Decentralized Collective Assault}
This subsection presents an example of the LMAS task, referred to as Decentralized Collective Assault (DCA).
Most current multi-agent environments are either designed under 
small-MAS setting with less than 30 agents or established in a simple discrete Gridworld \cite{zheng2018magent}.
In comparison, 
DCA aims to simulate a complex LMAS environment in continuous 2D space.
As shown in Figures~\ref{fig:dca}, 
blue tank agents team up against another script-controlled red team (with details given in Appendix).

The number of each team is denoted as $N_{blue}$ and $N_{red}$, 
and the number of total agents is adjusted in range $[100,300]$.
The goal is to survive and eliminate opponents.
A team wins when it wipes out the opponents or has more survivors when time runs out.
Limited by the scope of observation and random interference,
an agent $i$ only observes a nearby entity $j$ with probability $1-p_b$ when $\operatorname{dis}(j,i) \leq r_w$,
where $p_b \in [0,1)$ is the interference probability, 
and $\operatorname{dis}(\cdot)$ calculates the distance between agents.
The action space in the DCA environment is discrete with seven types of actions,
responsible for accelerating to four directions, 
rotating weapon clockwise, anti-clockwise, or doing nothing respectively.
Each agent has a fan-shaped weapon kill area with radius $\ell_i$ that eliminates opponents inside.
So far, the strength of a team largely depends on the initial agent number.
Thus we introduce map terrain that adds a new twist.
The terrain only affects the agent's fire radius $\ell_i$ by
$\ell_i=\ell_o  \cdot h_i$, 
where $\ell_o$ is a fix radius under flat terrain and 
$h_i \in (0,2)$ is the relative height factor,
which gives advantage to agents that have the high ground.
For the RL reward, each agent is rewarded $+1.0$ when make a kill, and $-0.5$ when get hit.

\subsection{Experimental Setup}
In experiments, we adjust the number of agents and level of interference in DCA.
Experiments with $5\%$ interference are denoted as $N_{blue}$-$N_{red}$-$S$,
and $10\%$ interference as $N_{blue}$-$N_{red}$-$L$.

We train our model with the PPO learner proposed in \cite{schulman2017proximal} and improved in \cite{ye2020mastering}.
In all experiments, the learning rate is $5\cdot 10^{-4}$, and the discount factor $\gamma$ is $0.99$.
At each update, we use trajectories collected from 64 episodes.
The GAE parameter $\lambda$ is $0.95$.
We select $d_c=2$ as default, 
and choose the Dual-ConcNet model shown in Fig.~\ref{new-cct-fig}(b) as an ablation baseline, referred to as Conc for simplicity.
The optional self-attention layer is \textbf{not} used unless referred to as Conc-SA.
Two ConcNet in this baseline model process \textit{friend-or-foe} entity observations respectively.
The experiments are performed with an RTX 8000 GPU, 
which takes around a day to train 50vs50 or 2 days to train 100vs100 from scratch.

We compare the default Conc model with plain MLP with zero-padding (PlainNet), 
soft self-attention (Soft-SA), as well as attention-based deep graph network (Att-DGN). 
(1) 
We construct a plain MLP policy without scalability and use zero-padding to maintain input dimensions,
this simple method is referred to as PlainNet.
The observation is converted to a fixed-length vector by simple concatenation followed by zero-padding.
(2)
We use soft SA to aggregate the observation sequence from entities.
This model shares similar attention structure used in \cite{hoshen2017vain} and \cite{iqbal2019actor}, 
and it is referred to as Soft-SA.
(3)
We implement a graph attention network that resembles \cite{agarwal2019learning} and \cite{deka2021natural}.
Unlike other models that strictly follow the paradigm of decentralized execution, 
the DGN-based model requires breaking the decentralization restrictions to work, and is referred to as Att-DGN.
(4)
We degenerate the main network of ConcNet into a special soft attention module, 
which only preserves top-$d_c$ attentional weights and zeros out other weights before performing softmax.
This Attention Rank-Out model (Conc-ARO) is designed to illustrate 
the non-trivial superiority of ConcNet compared to the soft-attention network.
(5)
We compare the baseline (Conc) with another Dual-ConcNet model (Conc-4Hist), 
which is designed for using history infomation and 
can use multi-step observations for decision making.

\begin{figure}[!t]
  \centering
  \begin{subfigure}[t]{0.45\linewidth}
  \centering
  \includegraphics[width=.999\linewidth]{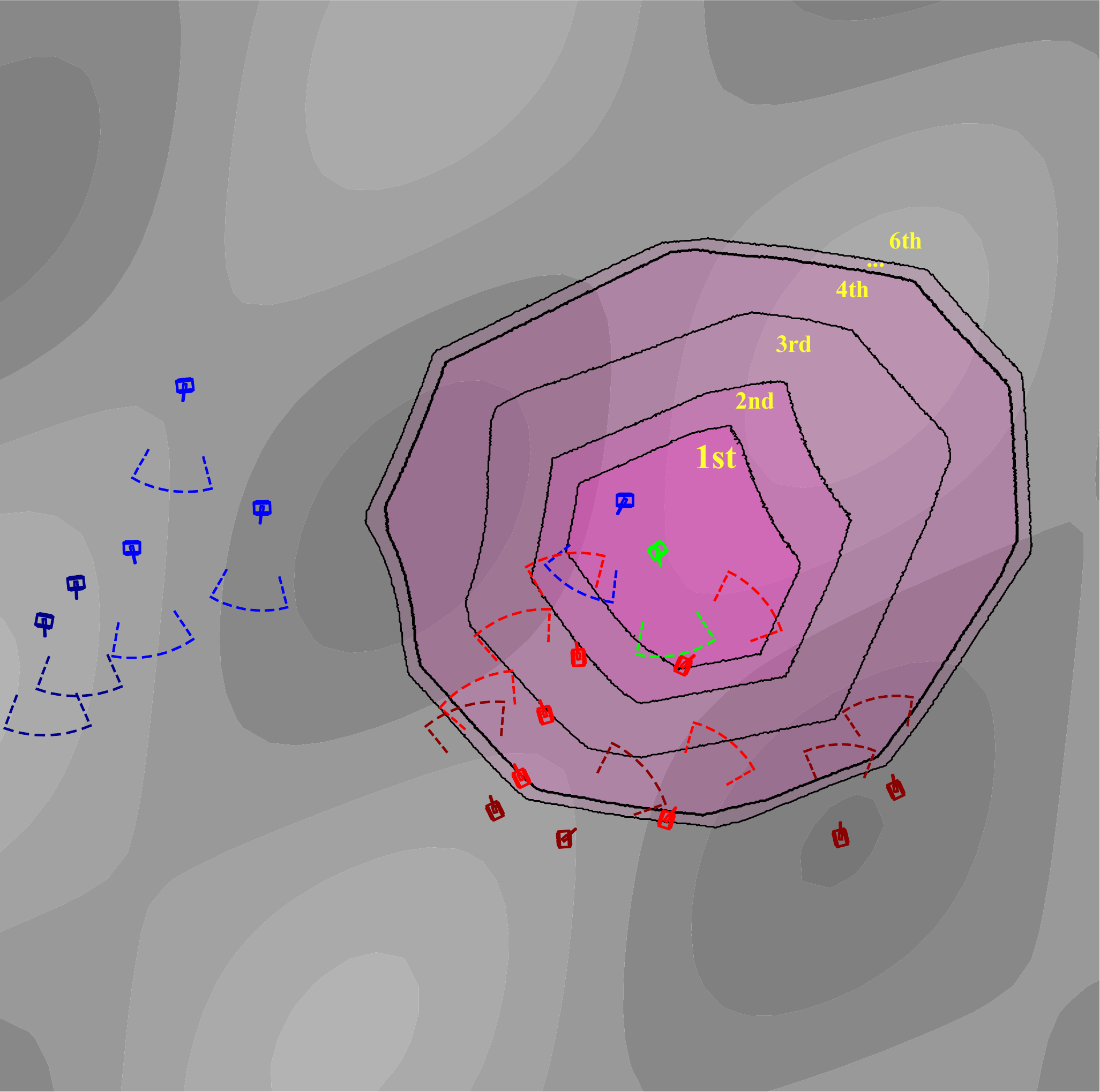}
  \caption{S1, rank of opponents.}
  \label{fig:dcadeepin1}
  \end{subfigure} 
\begin{subfigure}[t]{0.45\linewidth}
  \centering
  \includegraphics[width=.999\linewidth]{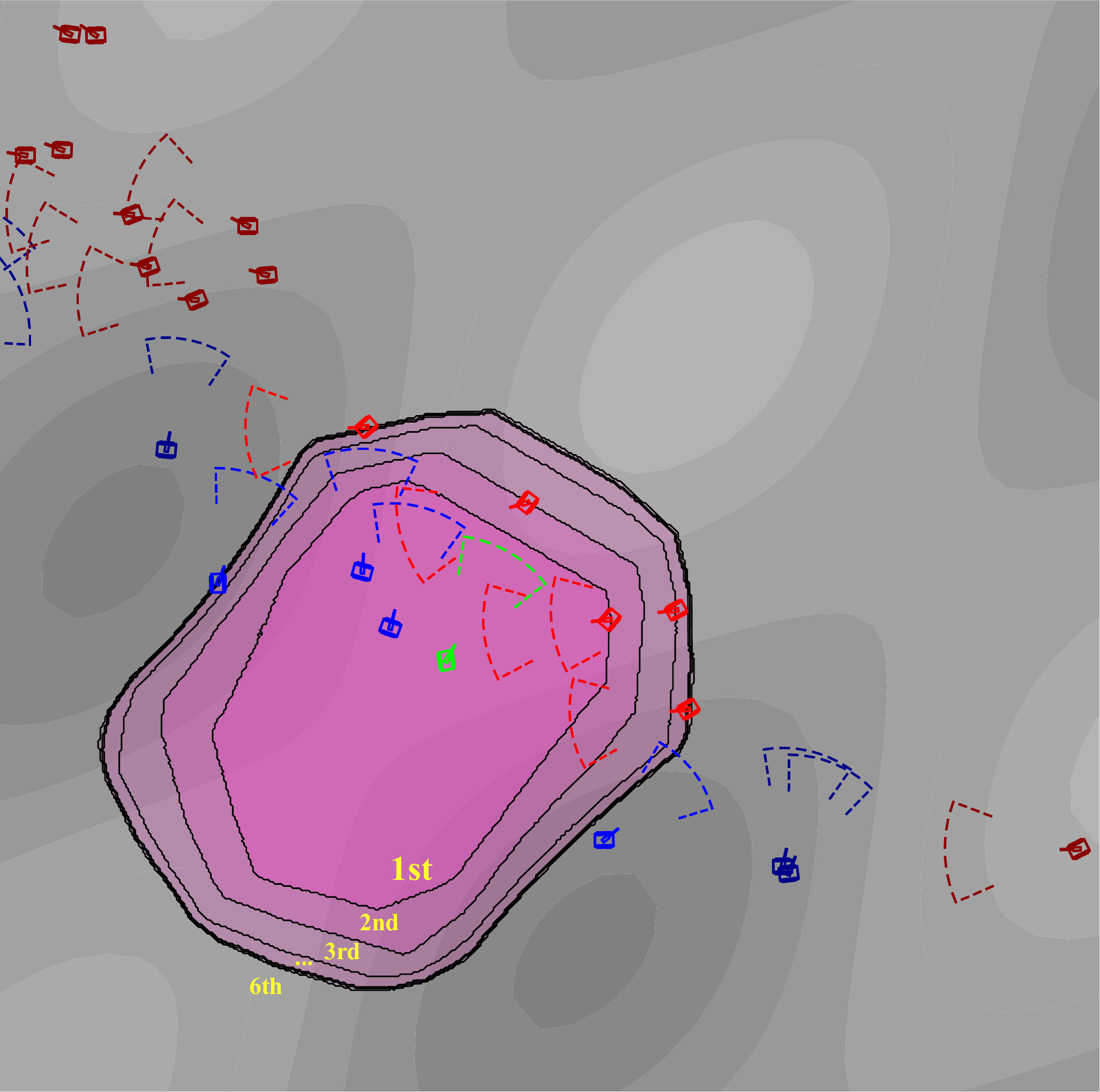}
  \caption{S2, rank of opponents.}
  \label{fig:dcadeepin2}
  \end{subfigure} 
  \\
\begin{subfigure}[t]{0.45\linewidth}
  \centering
  \includegraphics[width=.999\linewidth]{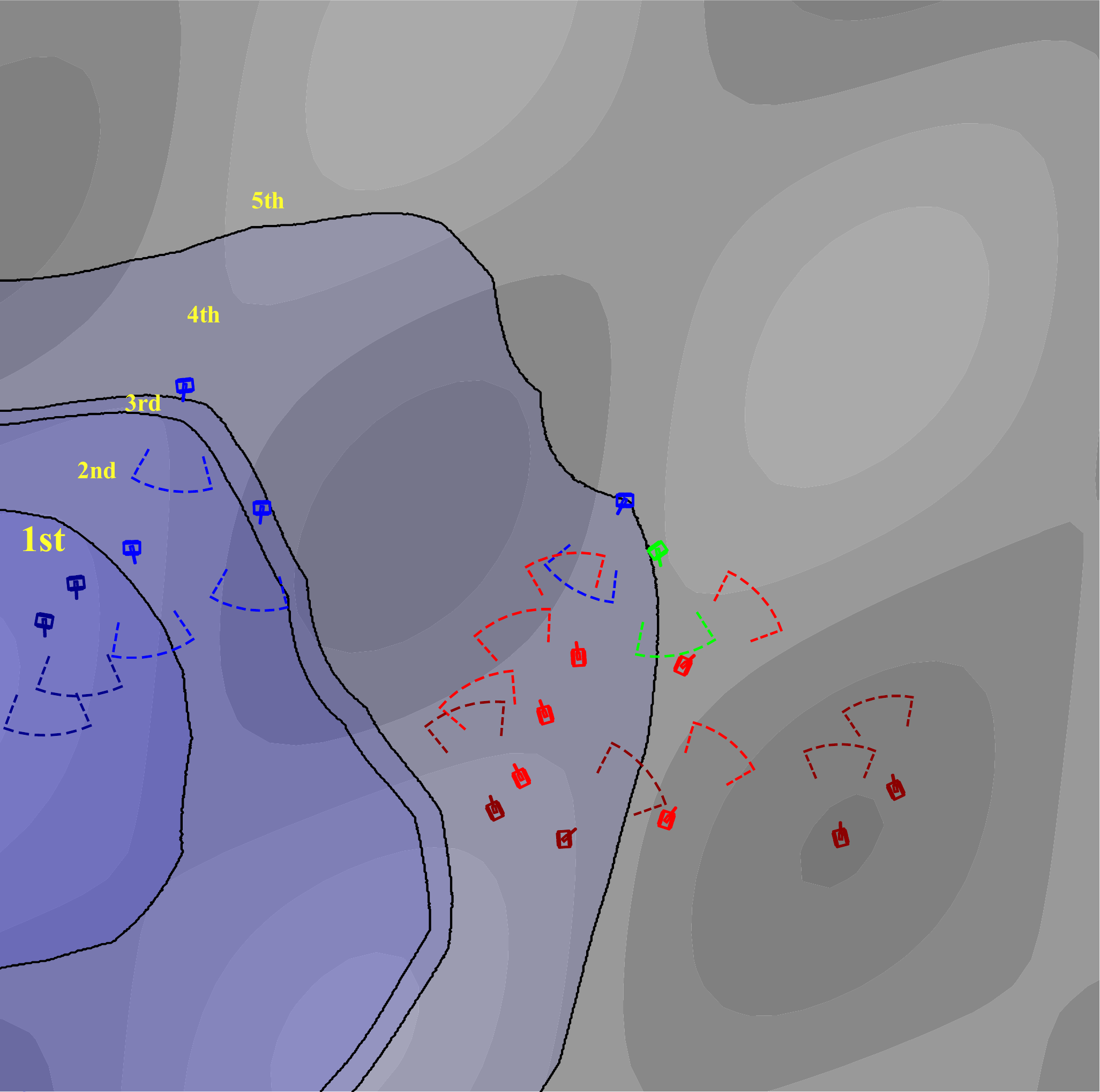}
  \caption{S1, rank of allies.}
  \label{fig:dcadeepin3}
  \end{subfigure} 
\begin{subfigure}[t]{0.45\linewidth}
  \centering
  \includegraphics[width=.999\linewidth]{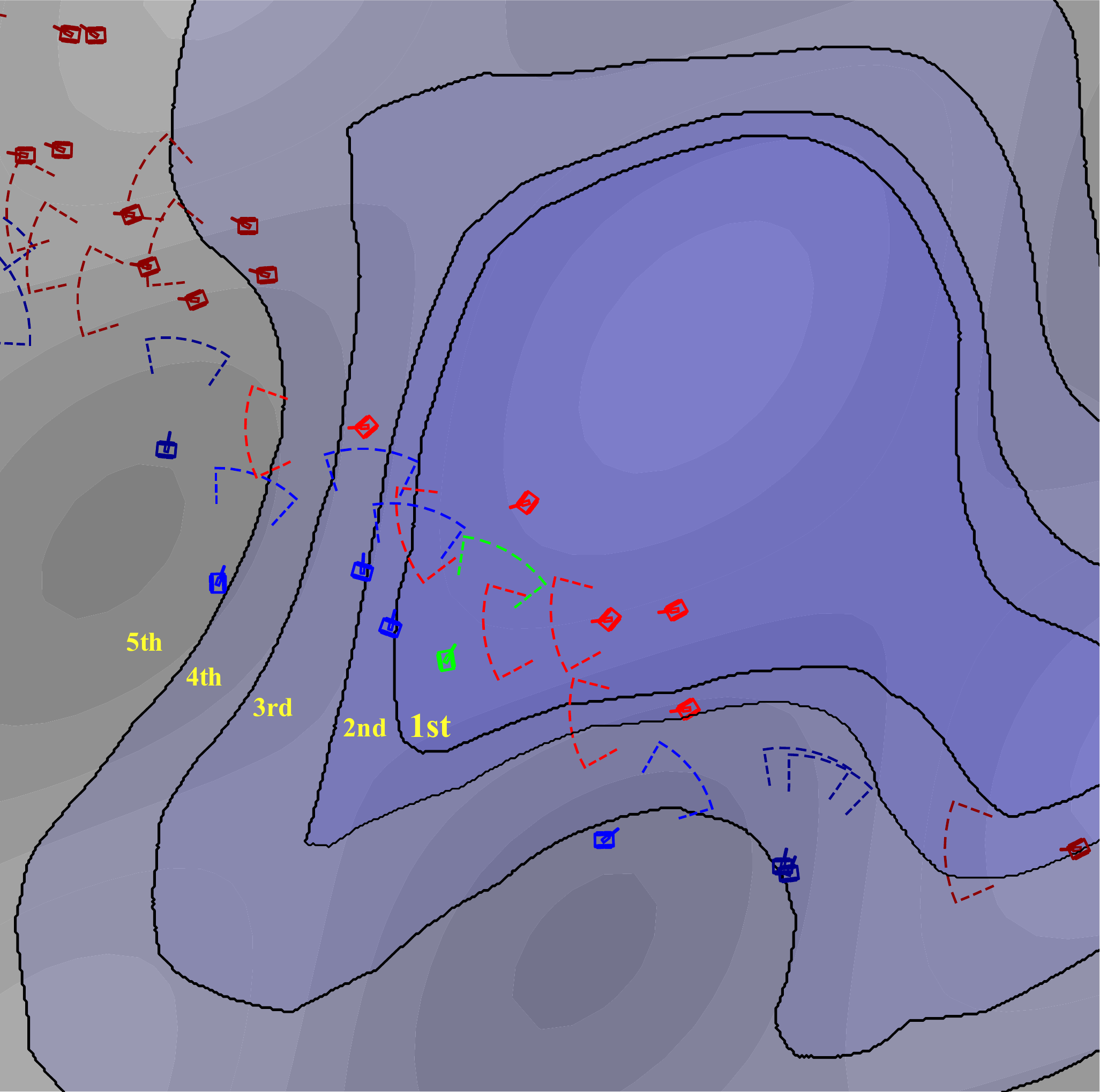}
  \caption{S2, rank of allies.}
  \label{fig:dcadeepin4}
  \end{subfigure} 

\caption{Visualizing the concentration score ranking under DCA.
We choose and freeze scenes 1 and 2 (S1 and S2) to investigate how the concentration network 
ranks the allies (blue) and opponents (red)
nearby an agent of interest (green). 
We use light-red and dark-red to indicate whether an opponent is visible to this agent of interest.
The terrain is represented by gray contours. 
The ranking is shown as pink and blue contours.
}
\label{fig:dcadeepin}
\end{figure}

\subsection{Ablations}
We perform a series of ablation experiments to answer the following questions.

\subsubsection{Scalability.}
We investigate how our method performs in even larger LMAS (Training-Scalability), 
and whether a trained model is robust to deal with scenarios with a different number of agents (Testing-Scalability).
We double the agent number from $50$ to $N_{blue}$=$N_{red}$=$100$,
while the training still begins from scratch under the same hyper-parameters.
Then the trained model is tested under different settings:
(a) Fixed number of RL agents $N_{blue}=100$, $N_{red}\in [50,150]$, 
(b) Fixed number of opponents $N_{red}=100$, $N_{blue}\in [50,150]$,
(c) Playing equal with $N_{blue}=N_{red} \in [50,150]$.

\subsubsection{Motivations.}
We examine whether both chosen motivational indices are essential.
Since our ConcNet is motivated by two motivational indices, 
we naturally doubt that one of the indices may not contribute to the model performance.
Thus we try to remove the survival time objective to inspect how it impacts the model.

\subsubsection{Details.}
(a) We assess how the concentration parameter $d_c$ influences the module.
(b) We investigate whether the extra self-attention layer shown in Fig.~\ref{new-cct-fig} provides improvement.
(c) We examine whether reducing our model from Dual-ConcNet to Single-ConcNet causes a performance decay.
(d) We investigate whether Conc-4Hist model can ultilize history observations under extreme interference.

It is necessary to reveal that none of these details are essential to the concentration network,
therefore showing the great flexibility of the concentration network.

\begin{figure}[t]
  \centering
  \includegraphics[width=1.0\linewidth]{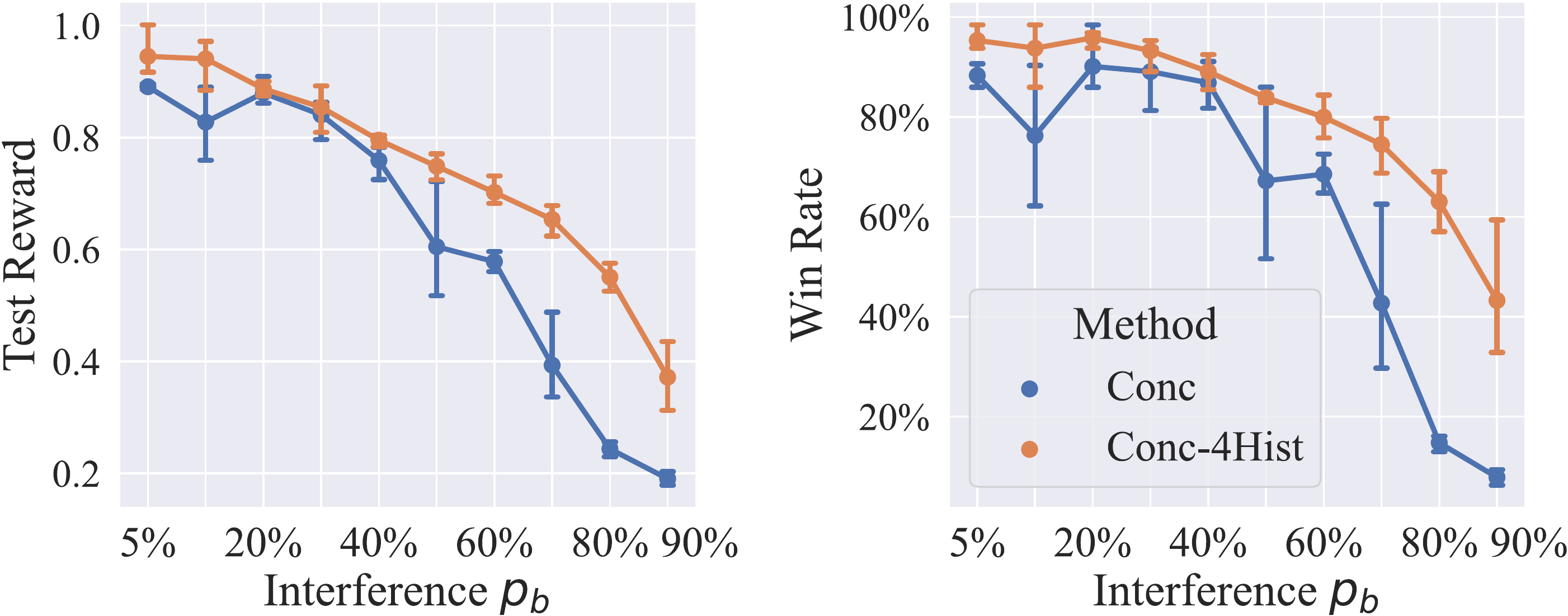}
  \caption{Testing Conc and Conc-4Hist models against different level of interference in 50vs50 DCA settings.
  In this test, the success probability of observing any entities $(1-p_b)$ reduces from 95\% to 10\%.}
  \label{fig:abl-hist}
\end{figure}

\section{Results}

Fig.~\ref{fig:comp} shows that concentration-based models significantly 
outperform other models.
In 50vs50-S settings, the average win rate of our model can reach 90\% while the win rate of other methods is below 30\%.

In 50vs50-L settings, the agent observations are blocked more frequently by interference.
Surprisingly, the increased interference benefits the win rate of attention-based methods.
This unforeseen result again indicates that the overwhelming observation is the bottleneck of soft attention models in LMAS,
since even random observation-blocking by interference can raise the performance of these models.
But after all, the win rate is still much lower than our concentration-based models.
Also, degenerated Conc-ARO model has weak performance against other Conc models, 
showing that simply imitating ConcNet with specially modified soft-attention does not have ideal results.

\subsubsection{Analyzing a Learned Concentration Network.}
It is possible to take a glimpse of what ConcNet has learned inside this black box.
In Fig.~\ref{fig:dcadeepin} we investigate how the concentration ranks a given entity in DCA.
We freeze two scenes and select an RL agent to investigate (plot as green), 
then we perform surgery on this agent's observation to insert a virtual entity nearby 
and test the concentration ranking.
Since we use Dual-ConcNet structure by default, 
we can study how the concentration network response to ally and opponent entities separately.

Fig.~\ref{fig:dcadeepin}(b) shows that besides its distance to opponents, 
an agent cares more about its rear since it is more vulnerable than its front.
The contour that ranks the ally concentration is significantly influenced by the terrain in Fig.~\ref{fig:dcadeepin}(d),
probably because the assistance of allies from the higher ground is more helpful in the future.
The ally concentration contour is unusual in Fig.~\ref{fig:dcadeepin}(c).
In this scene, the agent concentrates on faraway allies on the higher ground instead of its closest ally
since the reinforcements from faraway allies are more helpful to change the tide of the game.

\subsubsection{Scalability.}
By doubling the number of agents, the task difficulty increases significantly.
In the 100vs100(-S) experiment, the number of episodes required to train the policy doubles. 
Nevertheless, 
Fig.~\ref{100vs100} shows that our concentration-based models still 
has good performance under this setting, 
which demonstrates the Training-Scalability of the ConcNet model.

In order to evaluate the Testing-Scalability,
the trained model is tested under settings different in initial agent numbers in Fig.~\ref{fig:abl-scale}.
Starting from origin at $N_{blue}=N_{red}=100$.
The win rate drops slowly as opponents gradually outnumber the RL agents 
when the opponent number is increased, 
around -1\% per opponent.
In comparison, 
the win rate drops more rapidly when the number of RL agents is reduced, 
around -2\% per RL agent.
Alternatively, when the agent numbers of both sides change synchronously,
the trained policy adapts better since the game is fair in agent numbers.
Although the policy is trained under 100vs100, 
it still has about $50\%$ win rate in 130vs130 settings due to the Testing-Scalability. 
In Fig.~\ref{fig:abl-scale}(b), Two concentration network variants Conc and Conc-SA are compared.
In this setting, the default Conc model has a higher win rate without SA.

\subsubsection{Motivational Indices.}
Fig.~\ref{fig:abl-dc} shows that after removing the survival-time motivation index,
the performance is decayed considerably.
Without sufficient guidance of the essential motivation, 
the score function is unable to learn to rank the importance of entities.
Fortunately, these two motivational indices are general in almost all MAS tasks,
reflecting the underlying affinity between our concentration network and the LMAS policy gradient framework.
We emphasized that MAS problems from different research areas can be distinct from one another and need flexibility in solutions.
Our concentration network is open to any task-related motivational indices and can provide this flexibility.

\subsubsection{Structure Details.}
The parameter $d_c$ controls the number of entity representations to preserve in the pruning step.
Fig.~\ref{fig:abl-dc} shows that a larger $d_c$ slows down the learning process, 
but has no influence on the final reward after trained sufficiently.

The optional SA layer is helpful in 50vs50 settings, but at a great cost of GPU memory usage.
In comparison, Fig.~\ref{fig:abl-scale}(b) suggests that SA is detrimental to model performance under 100vs100.

Fig.~\ref{fig:abl-dc} also shows that Dual-ConcNet and Single-ConcNet have trivial performance differences. 
However, the Dual-ConcNet structure still has good reasons to be highlighted
because it shows the flexibility of the concentration network.
In heterogeneous MAS with many types of entities,
it is easy to extend Dual-ConcNet to Quad-Conc or even more complex structure to satisfy different requirements.

The multi-step version of Dual-ConcNet, Conc-4Hist, 
shows significant improvement when the interference level $p_b$ is extremely high.
According to Fig.~\ref{fig:abl-hist},
when the interference level is as high as $80\%$, 
agents of the Conc model suffer great disadvantage from going blind.
However, the Conc-4Hist model benefits from the ability to recall observations from history,
and shows significant improvement compared with the Conc model.

\section{Conclusions}

This paper aims at RL in Large-scale MAS (LMAS).
We start by modeling the process of concentration as a motivation-driven process,
and then put forward a concentration network specialized in processing
long sequences of entity observations in LMAS.
Furthermore, we propose a concentration policy gradient architecture that can train agent policies in LMAS from scratch.
Our concentration-based models not only significantly outperform existing MAS methods but also achieves excellent Training-Scalability as well as Testing-Scalability.
Moreover, 
we present and experimented with several variants of concentration policy gradient 
to demonstrate the flexibility of ConcNet.
Besides the two general motivational indices embedded in the concentration policy gradient,
Our concentration network is open to the implementation of any task-specific motivational indices 
to meet the requirements of distinct LMAS tasks.

For future work, 
we believe the concentration models can combine with transfer learning to learn more robust policies.
Moreover, 
we aim to apply the proposed concentration network to more multi-agent benchmark environments and 
investigate the possible applications in real-world problems.

\section{Acknowledgments}
This work was supported in part by 
the National Key Research and Development Program of China (2018AAA0102404), 
the National Natural Science Foundation of China (62073323), 
the Strategic Priority Research Program of Chinese Academy of Sciences (XDA27030403),
the External Cooperation Key Project of Chinese Academy Sciences (173211KYSB20200002),
and 
the Science and Technology Development Fund of Macau (No.0025/2019/AKP).

\clearpage
\appendix
\section*{Appendices}
\addcontentsline{toc}{section}{Appendices}
\renewcommand{\thesubsection}{\Alph{subsection}}

\setcounter{table}{0}   
\setcounter{figure}{0}
\renewcommand{\thetable}{A\arabic{table}}
\renewcommand{\thefigure}{A\arabic{figure}}

\subsection{Related Work}

Reinforcement learning has made remarkable achievements in single-agent 
problems such as Atari games.
It also holds promise for solving tasks in multi-agent systems.
Topics within multi-agent systems are diverse as there are many problems 
worth studying.
Most studies follow a paradigm referred to as centralized training with 
decentralized execution(CTDE).
In this paradigm, the agents can access anything during training but
are restricted from making independent decisions during execution.
MADDPG \cite{lowe2017multi} is one of the most classic methods 
that follow this paradigm.
In comparison, the Independently Q-Learning (IQL) \cite{watkins1989learning} 
isolates agent in both training and execution stages.
According to \cite{tampuu2017multiagent}, 
IQL is able to get good scores in some specific tasks 
but the non-stationary issue limits its performance in complex MAS environments.

A considerable amount of literature investigates sparse-reward tasks
and the credit assignment problem behind them.
One of the simplest methods is VDN \cite{sunehag2017value}, 
which assumes that the total Q-value is the sum of individual Q-values.
Qmix \cite{rashid2018qmix} points out that VDN is only able to 
represent a small part of all the possible Q-value decomposition
satisfying Individual-Global-Max (IGM), 
and uses hyper networks to construct an improved decomposition network.
Weighted Qmix \cite{rashid2020weighted}, and QTRAN \cite{son2019qtran} is then
proposed to provide even more general ways of decomposition.
Other studies use difficult approaches to achieve credit assignment.
COMA \cite{foerster2018counterfactual} takes advantage of CTDE and
estimates the counterfactual advantage function to determine the contribution of each agent.
QPD \cite{yang2020q} leverages the integrated gradient to directly decomposite Q-values.

Other research branches concentrate on learning more complex cooperative behavior 
to accomplish hard cooperative multi-agent tasks.
These studies usually weaken the constraints of CTDE and 
allow communication even in the execution stage.
In order to aggregate important features from an arbitrary number of agents,
soft attention module \cite{vaswani2017attention} is widely adopted.
In \cite{hoshen2017vain, iqbal2019actor}, 
a centralized critic based on attention module is put forward 
for more precise state value estimation.
Furthermore, 
it is found that the deep graph network \cite{scarselli2008graph} is an appropriate media to describe an environment involving multiple agents.
By defining the graph nodes as agents, and edges as agent interaction or communication,
studies \cite{agarwal2019learning, jiang2018graph, deka2021natural} reveal that
the DGN holds considerable promise for solving small-scale hard cooperative problems.
However, most existing studies are designed under small-team settings.
The issues of large-scale multi-agent systems are rarely studied.
In studies of NLP, 
it is found that soft attention has limitations dealing with long sentence sequence \cite{shen2018reinforced}.
When current attention-based RL algorithms are deployed in a rich-agent environment,
a similar performance degradation emerges due to the overwhelmed attention modules.

In the studies of Q-value decomposition, 
Starcraft micro-management environment \cite{vinyals2017starcraft, usunier2016episodic} is widely used,
currently, available maps support up to 27 agents in a team.
Multi-Agent Particle Environment (MAPE) \cite{lowe2017multi}, 
is a simple 2D multi-agent environment with great extensibility.
In most MAPE agent number is usually less than 10 \cite{iqbal2019actor}. 
Another interesting environment is hide-and-seek from \cite{baker2019emergent}, 
in which 2 hiders are trained to protect themselves from 2 seekers by using tools obtained from the environment.
The Fortattack environment used in \cite{deka2021natural} 
simulates a battle between 2 teams, each with 5 agents.
While there are various multi-agent test environments,
environments supporting at least a hundred agents are very rare.
The environment proposed in \cite{zheng2018magent} studies the competition of hundreds of agents. Nevertheless, it is studied in a discrete 2D grid environment and is oversimplified.
By consulting \cite{lowe2017multi,deka2021natural},
we develop a sophisticated battlefield simulation supporting up to 300 agents as the test environment 
called Decentralised Collective Assault (DCA).

In our work, we focus on the decentralized reinforcement learning problem for LMAS.
We put forward the concentration network specialized in LMAS tasks, referred to as ConcNet.
Furthermore, we propose a ConcNet-based policy gradient architecture capable of learning 
LMAS policy from scratch.





\begin{figure}[t]
  \centering
  \includegraphics[width=\linewidth]{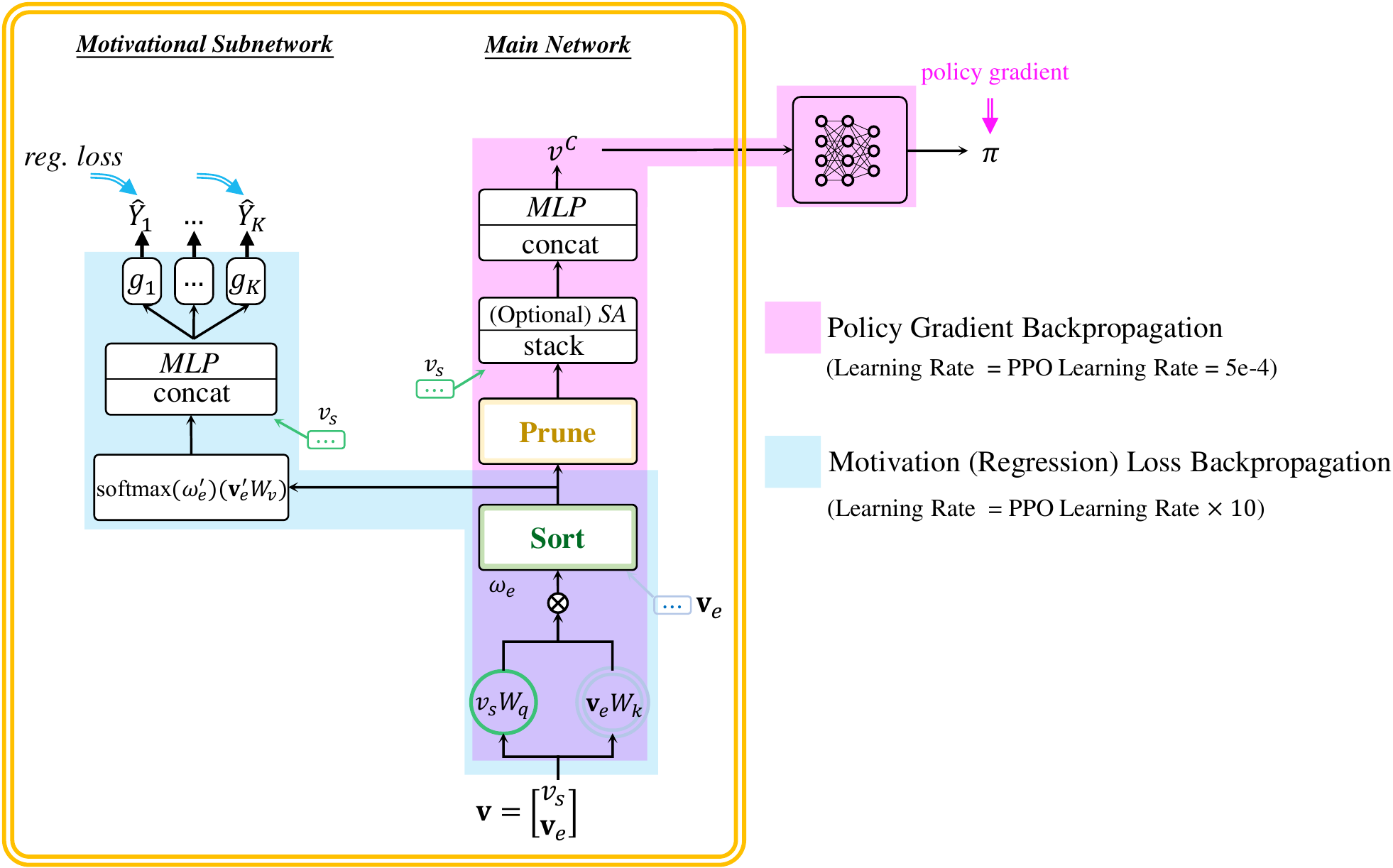}
  \caption{The gradient backpropagation path of ConcNet.}
  \label{fig:gradient}
\end{figure}

\begin{figure}[t]
  \centering
  \includegraphics[width=\linewidth]{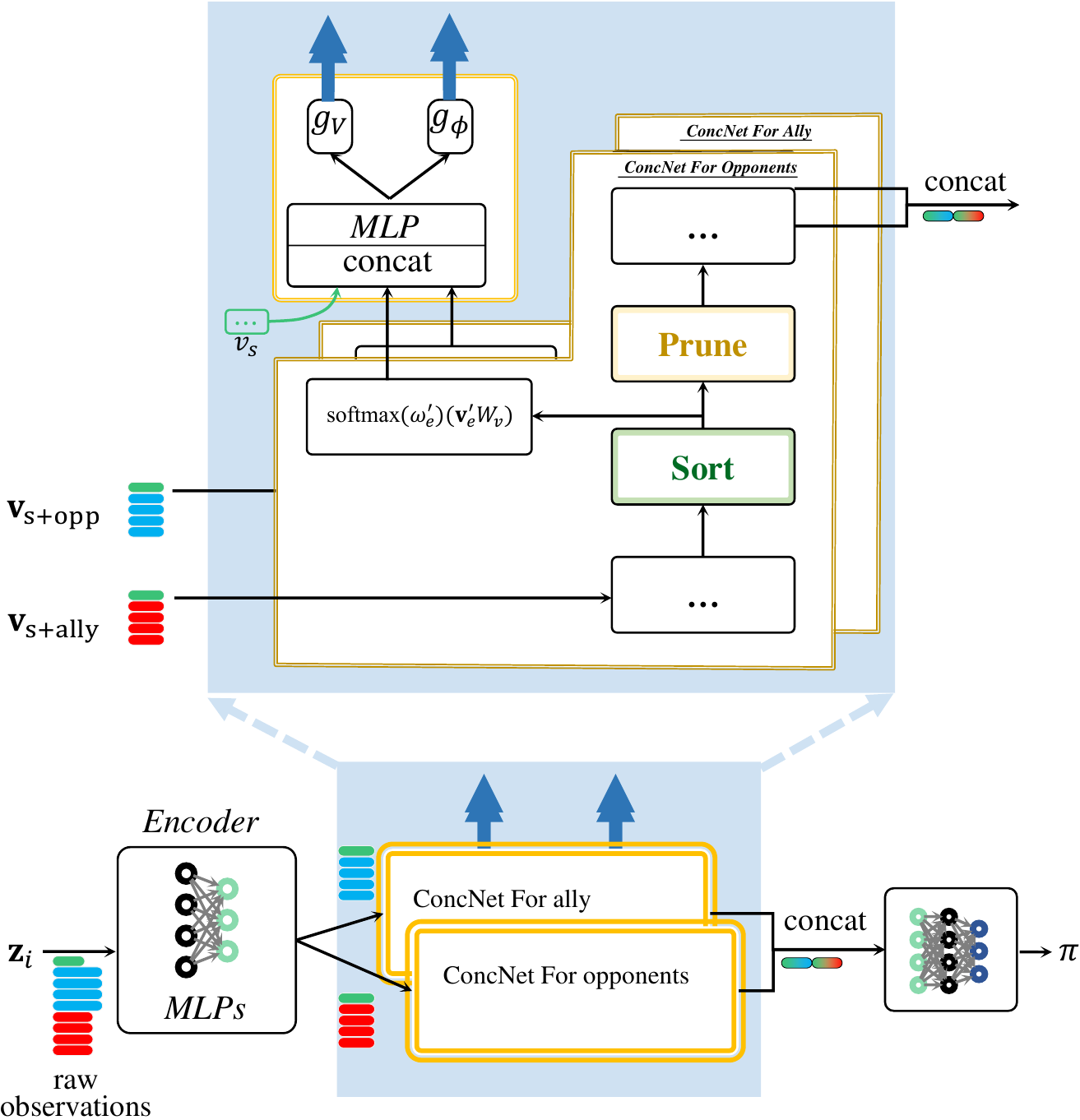}
  \caption{In the Dual-ConcNet model, two ConcNets produce only one set of motivational indices 
  because the motivational subnetworks are partially merged.}
  \label{fig:dualConcNet-merge}
\end{figure}

\subsection{Architecture Detail Explanation}
\subsubsection{Gradient Backpropagation}

\begin{figure}[t]
  \centering
  \includegraphics[width=\linewidth]{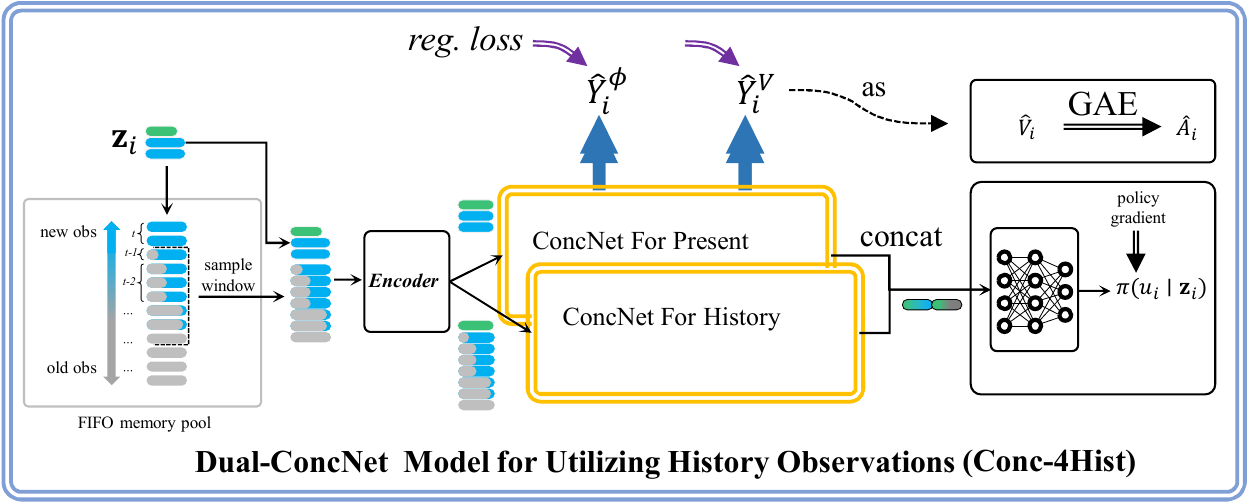}
  \caption{The architecture of Conc-4Hist.}
  \label{fig:conc-4hist}
\end{figure}
A complete ConcNet has a main network and a motivational subnetwork.
The backpropagation of the regression loss gradient and the policy gradient is shown in Fig.~\ref{fig:gradient}.
Different learning rates are applied to 
parameters trained by the motivation regression
and 
parameters trained only by policy gradient,
because the motivation regression is faster than the policy optimization.

\subsection{Architecture Optimization}
\begin{algorithm}[h]
    \caption{Concentration Policy Gradient}
    \label{alg:algorithm}
    \begin{algorithmic}[1] 
    \STATE Initialize all parameters $\theta=\theta_r \cup \theta_\pi$, empty buffer $\mathcal{D}$.
    \FOR {iteration = $1,2,\dots$}
        \STATE Run policy $\pi$ in environment for 64 episodes, load trajectories into $\mathcal{D}$.
        \STATE From $\mathcal{D}$, compute advantage estimation $\hat{A}_i(t)$ and calculate each motivational index $Y_k$.
        \FOR {epoch = $1$ to $N_{ppo}$}
            \STATE Compute regression gradient $\nabla_{\theta_r} \mathcal{L}_{reg}^{\theta_r}$ and 
            PPO policy gradient $\nabla_{\theta_\pi} J\left(\pi\right)$
            \STATE Perform regression gradient \textbf{descent} and policy gradient \textbf{ascent}, update parameters $\theta$.
        \ENDFOR
        \STATE empty buffer $\mathcal{D}$
    \ENDFOR
    \end{algorithmic}
    \end{algorithm}

\subsubsection{Merging Motivational Indices in Dual-ConcNet}

The Dual-ConcNet model is based on the simpler Single-ConcNet model,
and is designed for tasks with known friend-or-foe identification.
Two ConcNets are used to process ally entities and opponent entities separately.
However, it is not necessary to train upon twice the number of regression problems,
because two motivational subnetworks can merge before outputting the resulting indices, as shown in Fig.~\ref{fig:dualConcNet-merge}. 

\subsubsection{Utilizing History Observation with Conc-4Hist}

History observations are not always necessary depending on the tasks.
But when needed, our Conc model can easily support such history information utilization.
A simple approach is to insert an RNN at the tail of ConcNet,
but we are more interested in an alternative approach that 
can deeply integrate into the concentration policy gradient framework.
We present Conc-4Hist model as an example.

In this variant, 
we add a FIFO memory pool to store observations that an agent has experienced, as shown in Fig.~\ref{fig:conc-4hist}.
At each step, the memory pool accepts new observations and samples from history observation storage.
When a new episode starts, the memory pool must be cleared.
Then we re-route a Dual-ConcNet, which is originally designed to deal with entity observations of different types 
(e.g., \textit{friend-or-foe}), to process \textit{past-or-present} entity observations.
Note that Conc-4Hist also has to merge two motivational subnetworks as shown in Fig.~\ref{fig:dualConcNet-merge}.

\subsection{Optimization Details}
\subsubsection{Proximal Policy Optimization}
Policy gradient methods estimate the gradient of the policy parameters 
w.r.t. objective $J$ such as the discounted sum of rewards.
One of the most classic estimators of policy gradient is:
\begin{equation}
    \label{eq:pg-appendix}
    \nabla_{\theta} J\left(\pi\right)
    =
    \mathbb{E}
    \left[
          \nabla_{\theta} \log \pi_\theta \cdot \hat{A}_t
    \right]
\end{equation}
where $\hat{A}_{i}$ is the estimated advantage function.
PPO \cite{schulman2017proximal} is one of the most efficient
method in the family of policy optimization methods
and uses stochastic gradient ascent to perform each policy update.
PPO optimizes the objective $
\mathbb{E}\left[\min \left(l_{t}(\theta) \hat{A}_{t}, \operatorname{clip}\left(l_{t}(\theta), 1-\epsilon, 1+\epsilon\right) \hat{A}_{t}\right]\right.
$,
where $l_{t}(\theta)=\pi_\theta(u_t | s_t) / \pi_{old}(u_t | s_t)$ is 
the likelihood ratio between current policy and old policy.

\subsubsection{Hyperparameters}
\begin{table}[t]  
    \centering
    \begin{tabular}{ll}  
    \toprule   
    hyperparameter & value   \\  
    \midrule    
    discount factor $\gamma$ & 0.99   \\
    GAE  $\lambda$ & 0.95      \\
    Entropy coefficient   & 0.05      \\
    Gradient clipping  & 0.5      \\
    Number of episodes for each batch  & 64    \\
    PPO epochs $N_{ppo}$ & 24    \\
    Learning rate of policy gradient & $5\cdot 10^{-4}$      \\  
    Learning rate of regression & $5\cdot 10^{-3}$   \\    
    Motivation weight $\mu_1$ for $Y^{V}$ & 1.0 \\ 
    Motivation weight $\mu_2$ for $Y^{\phi}$ & 0.1 \\ 
    Threshold $T_{max}$ for $Y^{\phi}$& 10\\
    \bottomrule  
    \end{tabular}
    \caption{Hyperparameters in optimization}
    \label{tb:o-parameters}
\end{table}
We use following hyperparameter settings in optimization shown in Table.~\ref{tb:o-parameters},
and the network architecture follow hyperparameters in Table.~\ref{tb:n-parameters}.

\begin{table}[t]  
    \centering
    \begin{tabular}{ll}  
    \toprule   
    hyperparameter & value   \\  
    \midrule    
    Size of MLP layer $d_k$ & 48   \\
    Number of layers in each MLP & 1 or 2   \\
    Concentration hyperparameter $d_c$ & 2, 4 or 6   \\

    \bottomrule  
    \end{tabular}
    \caption{Network in optimization}
    \label{tb:n-parameters}
\end{table}

\subsection{DCA Environment Details}
\subsubsection{Observation.}
In DCA, each agent $i$ is able to observe a nearby entity $j$ (ally agent or opponent agent)
with success probability $p_{ij}=1-p_b$ at each time step.
If the observation is successful,
the agent can obtain the position, speed, weapon direction, terrain, and team belonging of $j$.

Note that only the RL team is limited by observation.
The build-in opponent AI (red team) has access to states of everything.
Thus, the probability of interference $p_{ij}$ is effective only on RL algorithms.

\subsubsection{Action.}
The action space in the DCA environment is discrete with seven types of action signals,
responsible for staying idle, accelerating to $(+x,-x,+y,-y)$, rotating weapon clockwise or anti-clockwise, respectively.
To create difficulty, 
RL (blue) agents have a limited weapon rotation speed of 10 degrees per game step,
while the build-in opponent AI does not suffer from this limitation
and can attack agents at its rear by instantly rotating weapon.

\subsubsection{Terrain}
The map terrain in DCA influences an agent's fire radius $\ell_i$ by $\ell_i=\ell_o  \cdot h_i$,
where $\ell_o$ is a fix radius under flat terrain and 
$h_i \in (0,2)$ is the relative height factor.
We use a Rastrigrin-like function to shape the terrain:
\begin{equation}
    \begin{aligned}
        h_i = 1 &+\lambda_A\left[ \cos(3\pi x_i/5) + \cos(\pi y_i)\right] \\
                &- \lambda_B\left[(x_i/10)^2+(y_i/10)^2\right] 
        \label{eq.terrain}
    \end{aligned}
\end{equation}
where $(x_i,y_i)$ is the position of the agent.
The parameter $\lambda_A$ adjusts the undulation of terrain, 
which expands the advantage that an agent possesses by
taking control of higher ground before its opponents.
And larger $\lambda_B$ discourages agents from leaving the center area of the map, which is slightly higher. 
In experiment we use $\lambda_A=0.05$ and $\lambda_B= 0.2$.

\subsubsection{Opponent AI against RL agents.}
\label{appendix:RedAI}
The opponent red team utilized a policy written in a built-in script controller with fixed rules to train robust RL agents.
The opponent team has cheating advantages
because it has access to all environment states,
considers no communication limitation.
And most importantly,
agents in the opponent team are not limited by weapon rotation speed and can instantly rotate the weapon to attack enemies approaching from any direction.
We create a simple but effective multi-agent policy with situation assessment capability using an method inspired by the theory of military operations. 
The agents will form several teams based on the distribution of enemies and coordinated attacks on the nearest agent.
The basic idea of the opponent controller includes:
1) Creating virtual groups with k-means clustering by $L_2$ distance for both red and blue teams.
2) Associating each red group with a target opponent group.
3) Assigning every group agent with an opponent from its target group using the Hungarian algorithm.

\begin{figure}[tb]
  \centering
  \includegraphics[width=\linewidth]{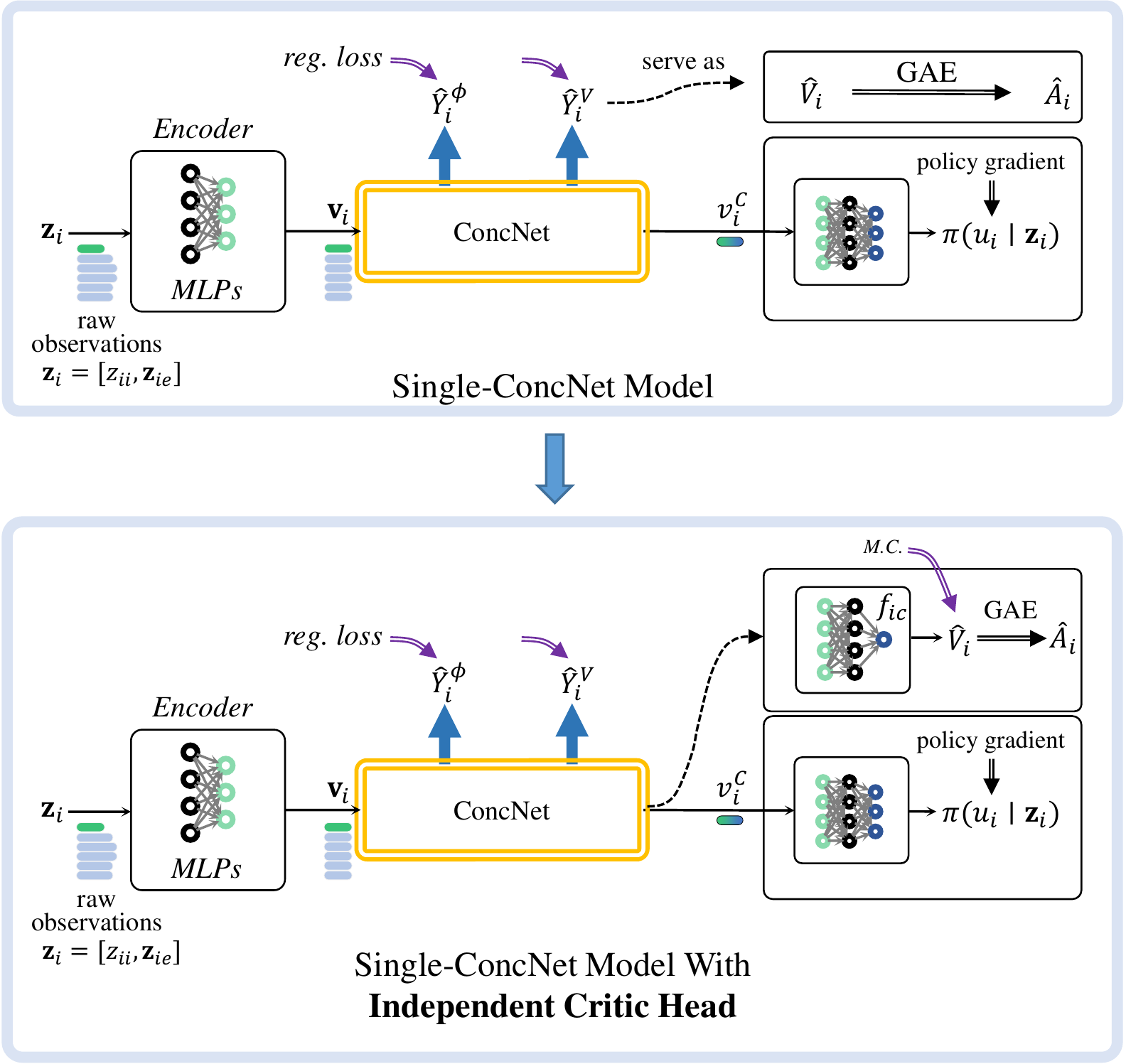}
  \caption{The structure of the concentration policy gradient with independent critic head.
  The Single-ConcNet model is used to demonstrate the difference.}
  \label{fig:ind-critic-structure}
\end{figure}

\begin{figure}[tb]
  \centering
  \includegraphics[width=\linewidth]{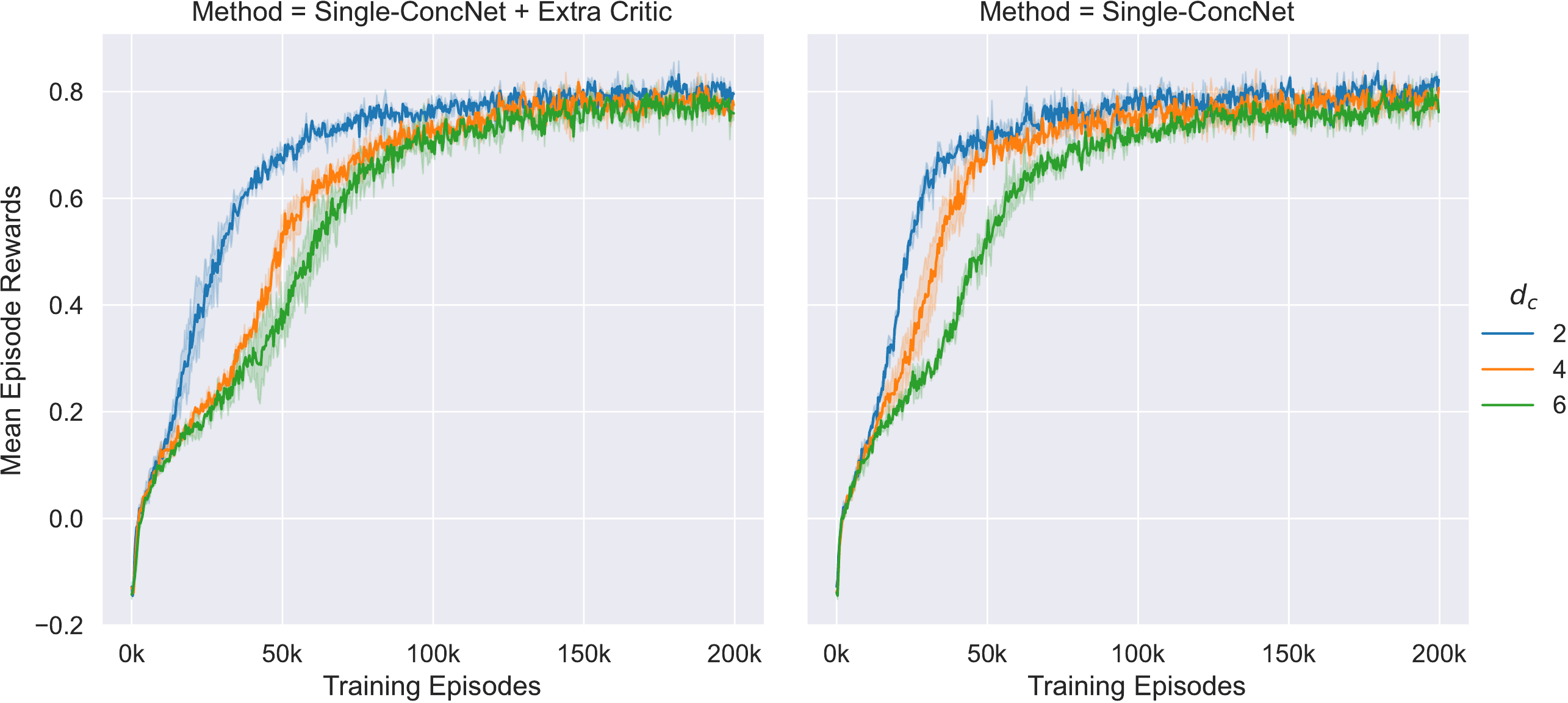}
  \caption{The comparison between Single-Conc and its variant with independent critic head.}
  \label{fig:ind-critic}
\end{figure}
\begin{figure*}[tb]
\centering
  \begin{subfigure}[t]{0.28\linewidth}
    \centering
    \includegraphics[width=\linewidth]{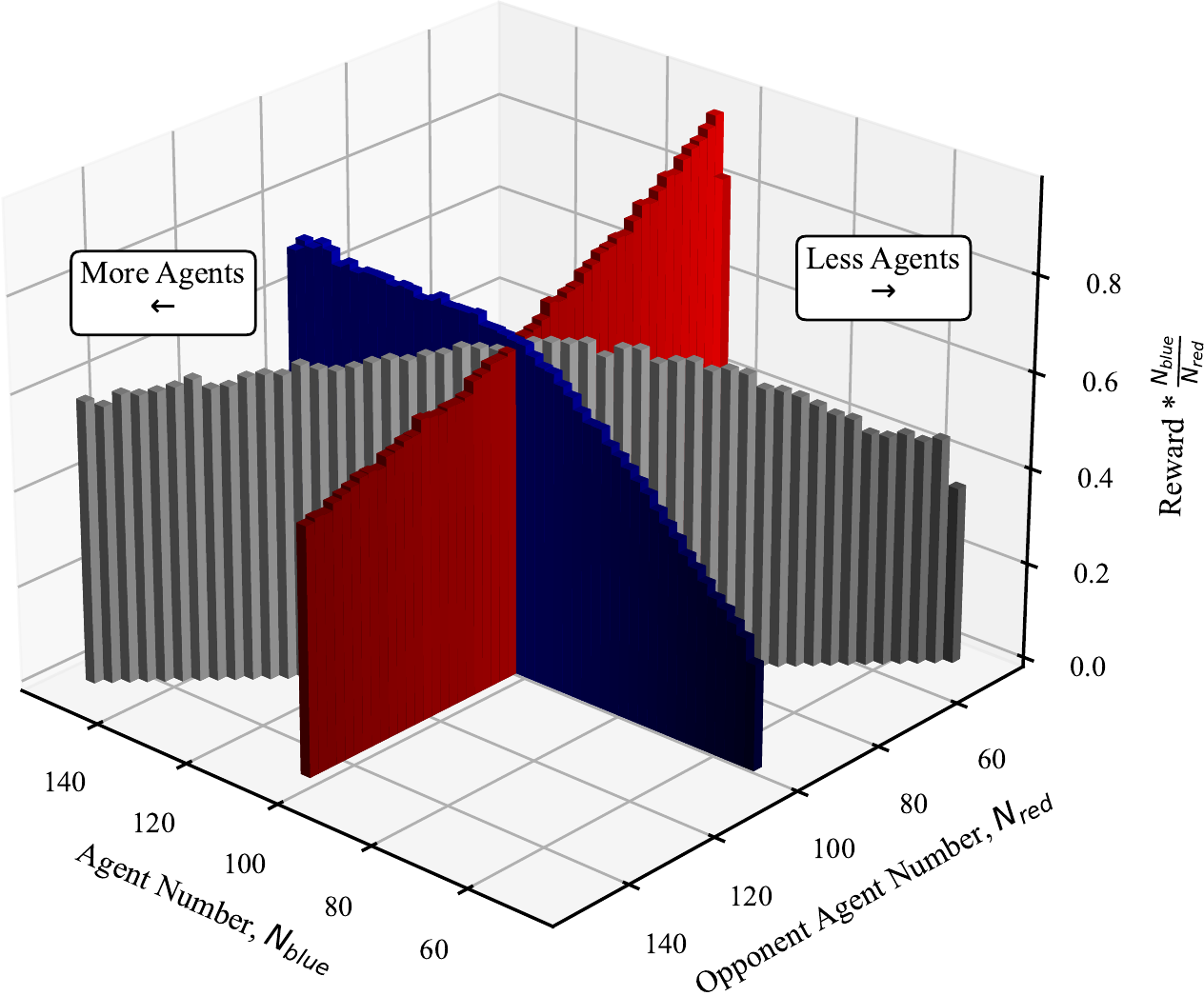}
    \caption{Average Test Reward of Conc}
    \end{subfigure} 
  \begin{subfigure}[t]{0.28\linewidth}
    \centering
    \includegraphics[width=\linewidth]{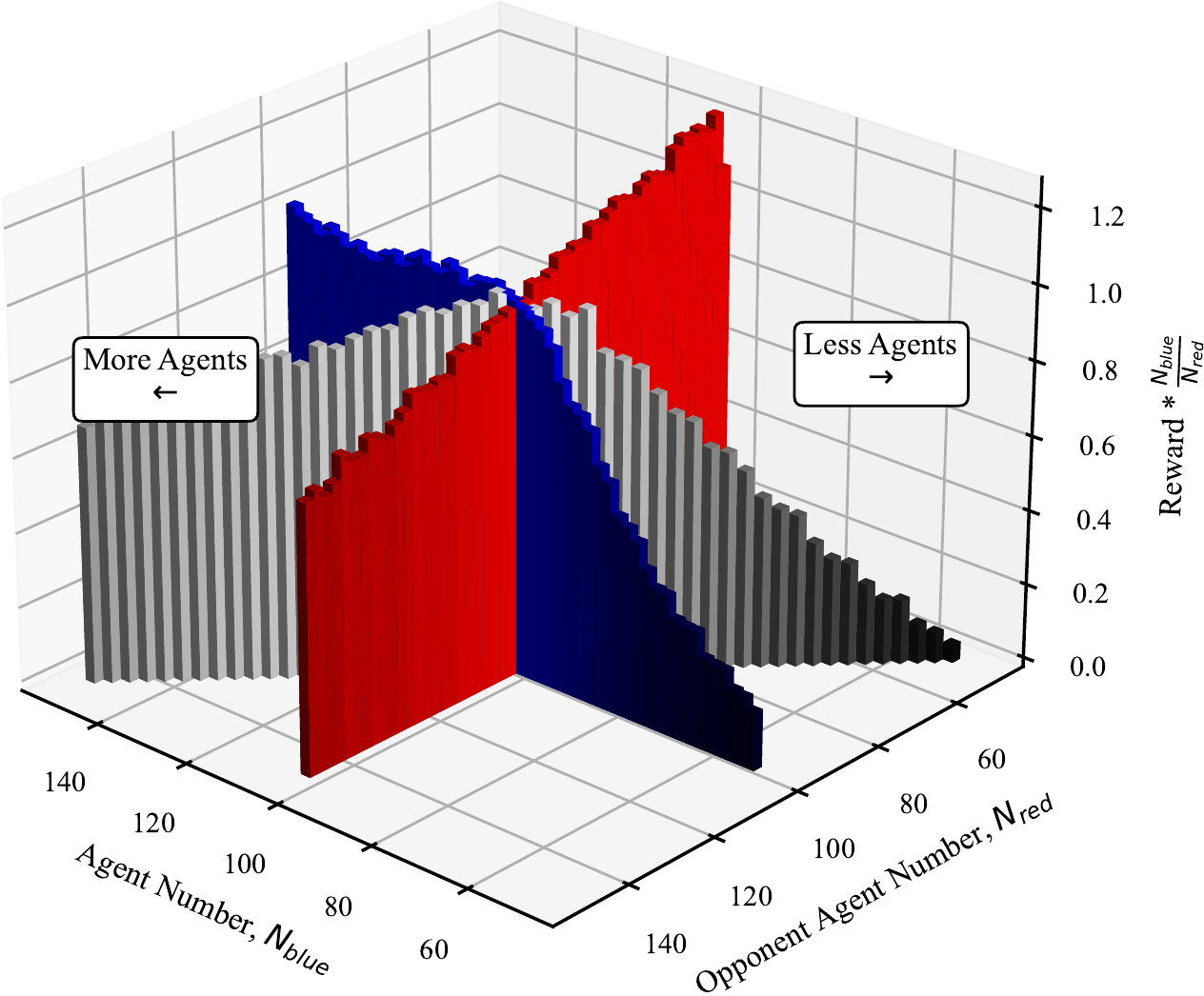}
    \caption{Average Test Reward of Conc-SA}
    \end{subfigure} 
    \\
  \begin{subfigure}[t]{0.28\linewidth}
    \centering
    \includegraphics[width=\linewidth]{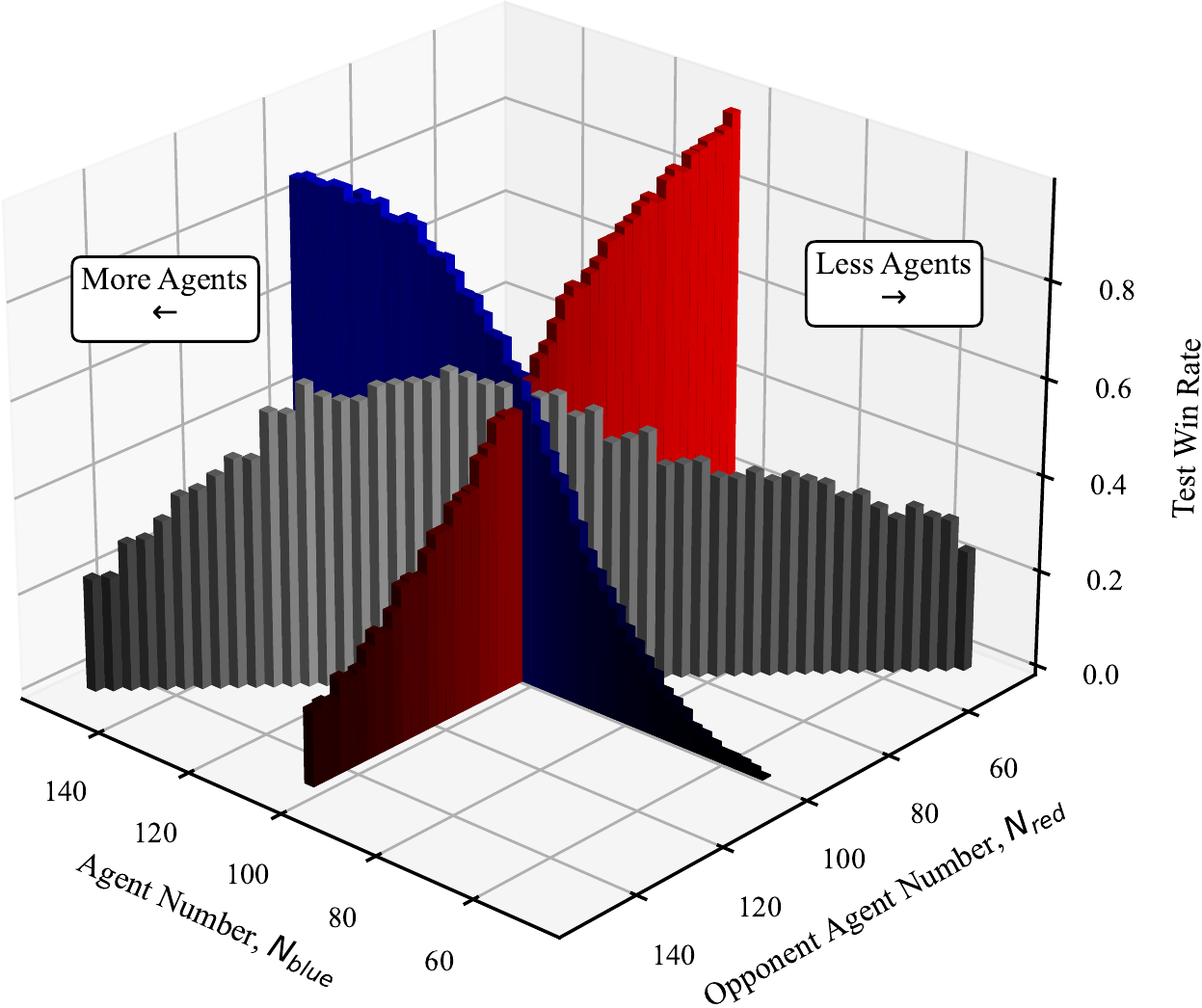}
    \caption{Test Win Rate of Conc}
    \end{subfigure} 
  \begin{subfigure}[t]{0.28\linewidth}
    \centering
    \includegraphics[width=\linewidth]{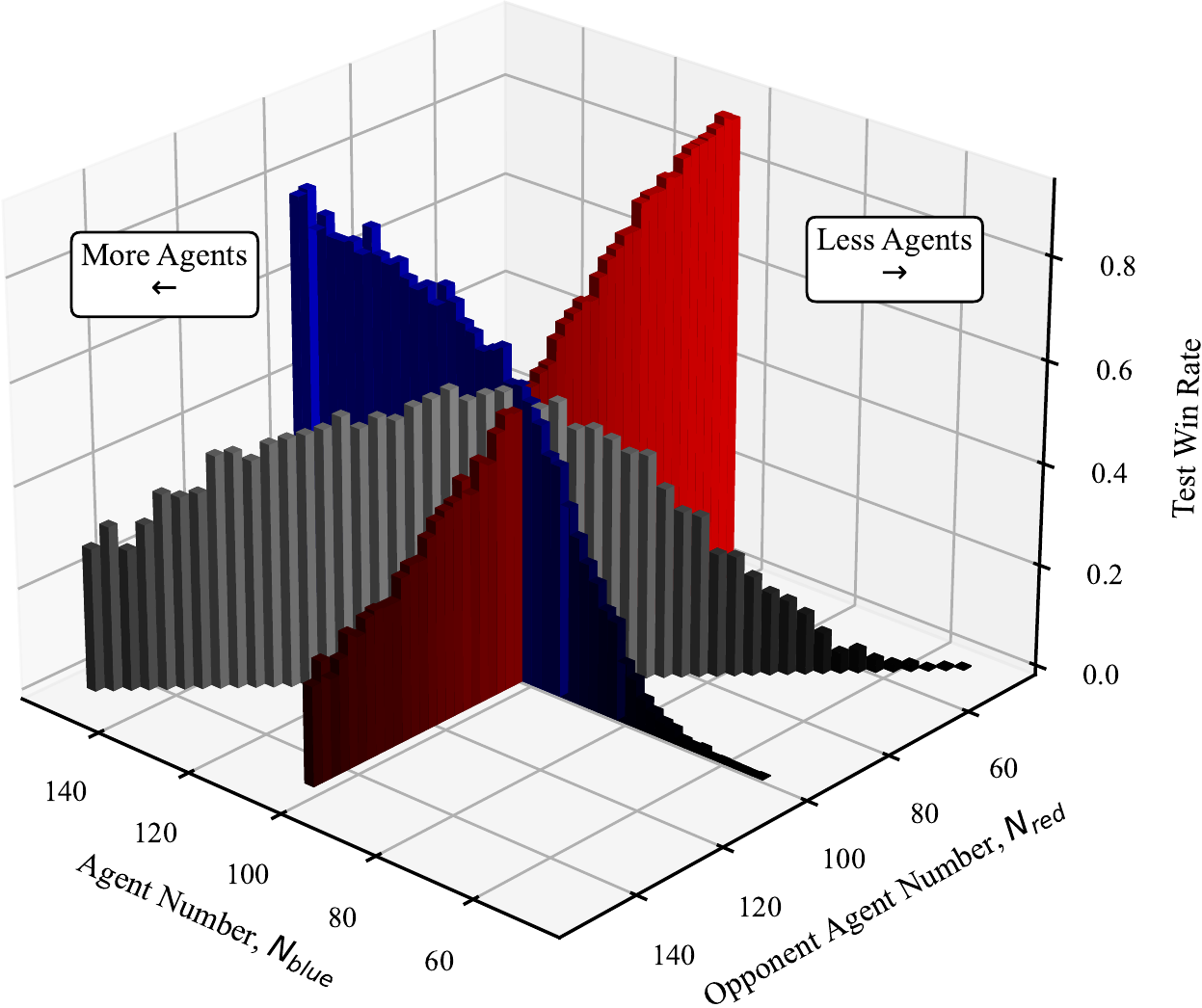}
    \caption{Test Win Rate of Conc-SA}
    \end{subfigure} 

  \caption{Using reward and win-rate to evaluate Test-Scalability. }
  \label{fig:test-scalability}
\end{figure*}

\begin{figure*}[th]
    \centering
    \includegraphics[width=0.75\linewidth]{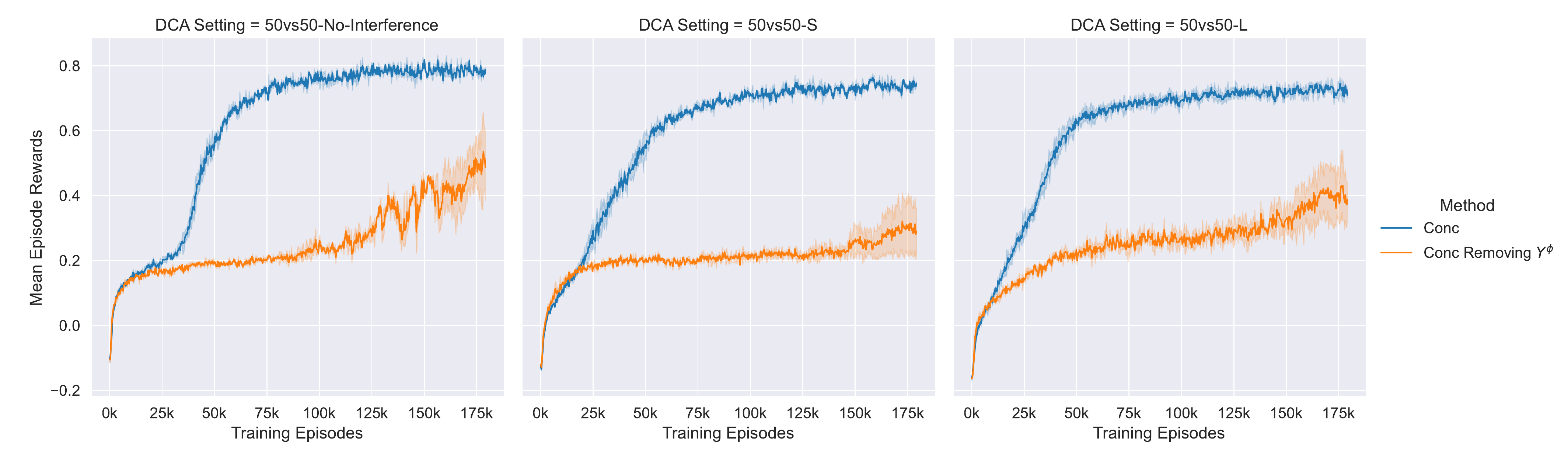}
    \caption{The removing one of the survival-time motivational indices under Different Interference Level.}
    \label{fig:rm-motiv2}
  \end{figure*}
\subsection{Other Experiments}
\subsubsection{Concentration Policy Gradient Architecture with Independent Critic Head}
In addition to the structure shown in Fig.~\ref{new-cct-fig}, 
we also investigate another possible model structure that 
utilizes the output of ConcNet's main network to estimate the state value,
instead of directly using one of the motivation indices as the state value.
Note that in this variant model,
the calculations of the value function and motivation indices
are isolated and need to be trained independently.
As shown in Fig.~\ref{fig:ind-critic-structure},
the state value is now estimated by $f_{ic}(\cdot)$:
$$
    \hat{V}_i = f_{ic}(v_i^c)
$$
We use the Monte Carlo method to train the parameters of $f_{ic}(\cdot)$
and network parameters that produced $v_i^c$.
Next, this variant is compared with the original Single-ConcNet model,
and the result is shown in Fig.~\ref{fig:ind-critic}.
It is illustrated in Fig.~\ref{fig:ind-critic} that 
introducing an independent critic head has no significant influence 
on the model performance.

\subsubsection{Ablation of Motivational Indices under Different Interference Level}
In the previous experiment shown in Fig.~\ref{fig:abl-dc},
we investigate the influence of removing the survival-time motivation index.
We perform the same tests under different levels of DCA interference.
As shown in Fig.~\ref{fig:rm-motiv2}, removing the survival-time motivational index
leads to more negative impacts when stronger interference exists.

\subsubsection{Test-Scalability Experiments}
We use reward and win-rate as metrics to evaluate test-scalability in Fig.~\ref{fig:test-scalability}.
The two models, Conc and Conc-SA, is selected and trained in the 100vs100-S setting.
Next, they are tested under settings with different number of agents.
For $N_{blue} \neq N_{red}$, we process the average test reward with $r'=r\cdot\frac{N_{blue}}{N_{red}}$, 
as shown in Fig.~\ref{fig:test-scalability}

\bibliography{cite.bib}

\begin{thebibliography}{37}
\providecommand{\natexlab}[1]{#1}

\bibitem[{Agarwal, Kumar, and Sycara(2019)}]{agarwal2019learning}
Agarwal, A.; Kumar, S.; and Sycara, K. 2019.
\newblock Learning transferable cooperative behavior in multi-agent teams.
\newblock \emph{arXiv preprint arXiv:1906.01202}.

\bibitem[{Astle and Scerif(2009)}]{astle2009using}
Astle, D.~E.; and Scerif, G. 2009.
\newblock Using developmental cognitive neuroscience to study behavioral and
  attentional control.
\newblock \emph{Developmental Psychobiology: The Journal of the International
  Society for Developmental Psychobiology}, 51(2): 107--118.

\bibitem[{Baker et~al.(2019)Baker, Kanitscheider, Markov, Wu, Powell, McGrew,
  and Mordatch}]{baker2019emergent}
Baker, B.; Kanitscheider, I.; Markov, T.; Wu, Y.; Powell, G.; McGrew, B.; and
  Mordatch, I. 2019.
\newblock Emergent tool use from multi-agent autocurricula.
\newblock \emph{arXiv preprint arXiv:1909.07528}.

\bibitem[{Chen et~al.(2019)Chen, Liu, Kreiss, and Alahi}]{chen2019crowd}
Chen, C.; Liu, Y.; Kreiss, S.; and Alahi, A. 2019.
\newblock Crowd-robot interaction: Crowd-aware robot navigation with
  attention-based deep reinforcement learning.
\newblock In \emph{2019 International Conference on Robotics and Automation
  (ICRA)}, 6015--6022. IEEE.

\bibitem[{Deka and Sycara(2021)}]{deka2021natural}
Deka, A.; and Sycara, K. 2021.
\newblock Natural emergence of heterogeneous strategies in artificially
  intelligent competitive teams.
\newblock In \emph{International Conference on Swarm Intelligence}, 13--25.
  Springer.

\bibitem[{Diallo and Sugawara(2020)}]{diallo2020multi}
Diallo, E. A.~O.; and Sugawara, T. 2020.
\newblock Multi-Agent Pattern Formation: a Distributed Model-Free Deep
  Reinforcement Learning Approach.
\newblock In \emph{2020 International Joint Conference on Neural Networks
  (IJCNN)}, 1--8. IEEE.

\bibitem[{Foerster et~al.(2018)Foerster, Farquhar, Afouras, Nardelli, and
  Whiteson}]{foerster2018counterfactual}
Foerster, J.; Farquhar, G.; Afouras, T.; Nardelli, N.; and Whiteson, S. 2018.
\newblock Counterfactual multi-agent policy gradients.
\newblock In \emph{Proceedings of the AAAI Conference on Artificial
  Intelligence}, volume~32.

\bibitem[{Hoshen(2017)}]{hoshen2017vain}
Hoshen, Y. 2017.
\newblock Vain: Attentional multi-agent predictive modeling.
\newblock \emph{arXiv preprint arXiv:1706.06122}.

\bibitem[{Iqbal and Sha(2019)}]{iqbal2019actor}
Iqbal, S.; and Sha, F. 2019.
\newblock Actor-attention-critic for multi-agent reinforcement learning.
\newblock In \emph{International Conference on Machine Learning}, 2961--2970.
  PMLR.

\bibitem[{Jiang et~al.(2018)Jiang, Dun, Huang, and Lu}]{jiang2018graph}
Jiang, J.; Dun, C.; Huang, T.; and Lu, Z. 2018.
\newblock Graph convolutional reinforcement learning.
\newblock \emph{arXiv preprint arXiv:1810.09202}.

\bibitem[{Kraemer and Banerjee(2016)}]{kraemer2016multi}
Kraemer, L.; and Banerjee, B. 2016.
\newblock Multi-agent reinforcement learning as a rehearsal for decentralized
  planning.
\newblock \emph{Neurocomputing}, 190: 82--94.

\bibitem[{Lowe et~al.(2017)Lowe, Wu, Tamar, Harb, Abbeel, and
  Mordatch}]{lowe2017multi}
Lowe, R.; Wu, Y.; Tamar, A.; Harb, J.; Abbeel, P.; and Mordatch, I. 2017.
\newblock Multi-agent actor-critic for mixed cooperative-competitive
  environments.
\newblock \emph{arXiv preprint arXiv:1706.02275}.

\bibitem[{Neishi and Yoshinaga(2019)}]{neishi2019relation}
Neishi, M.; and Yoshinaga, N. 2019.
\newblock On the relation between position information and sentence length in
  neural machine translation.
\newblock In \emph{Proceedings of the 23rd Conference on Computational Natural
  Language Learning (CoNLL)}, 328--338.

\bibitem[{Oliehoek and Amato(2016)}]{oliehoek2016concise}
Oliehoek, F.~A.; and Amato, C. 2016.
\newblock \emph{A concise introduction to decentralized POMDPs}.
\newblock Springer.

\bibitem[{Oliehoek, Spaan, and Vlassis(2008)}]{oliehoek2008optimal}
Oliehoek, F.~A.; Spaan, M.~T.; and Vlassis, N. 2008.
\newblock Optimal and approximate Q-value functions for decentralized POMDPs.
\newblock \emph{Journal of Artificial Intelligence Research}, 32: 289--353.

\bibitem[{Rashid et~al.(2020)Rashid, Farquhar, Peng, and
  Whiteson}]{rashid2020weighted}
Rashid, T.; Farquhar, G.; Peng, B.; and Whiteson, S. 2020.
\newblock Weighted qmix: Expanding monotonic value function factorisation.
\newblock \emph{arXiv e-prints}, arXiv--2006.

\bibitem[{Rashid et~al.(2018)Rashid, Samvelyan, Schroeder, Farquhar, Foerster,
  and Whiteson}]{rashid2018qmix}
Rashid, T.; Samvelyan, M.; Schroeder, C.; Farquhar, G.; Foerster, J.; and
  Whiteson, S. 2018.
\newblock Qmix: Monotonic value function factorisation for deep multi-agent
  reinforcement learning.
\newblock In \emph{International Conference on Machine Learning}, 4295--4304.
  PMLR.

\bibitem[{Rubenstein, Cornejo, and Nagpal(2014)}]{rubenstein2014programmable}
Rubenstein, M.; Cornejo, A.; and Nagpal, R. 2014.
\newblock Programmable self-assembly in a thousand-robot swarm.
\newblock \emph{Science}, 345(6198): 795--799.

\bibitem[{Scarselli et~al.(2008)Scarselli, Gori, Tsoi, Hagenbuchner, and
  Monfardini}]{scarselli2008graph}
Scarselli, F.; Gori, M.; Tsoi, A.~C.; Hagenbuchner, M.; and Monfardini, G.
  2008.
\newblock The graph neural network model.
\newblock \emph{IEEE transactions on neural networks}, 20(1): 61--80.

\bibitem[{Schulman et~al.(2015)Schulman, Moritz, Levine, Jordan, and
  Abbeel}]{schulman2015high}
Schulman, J.; Moritz, P.; Levine, S.; Jordan, M.; and Abbeel, P. 2015.
\newblock High-dimensional continuous control using generalized advantage
  estimation.
\newblock \emph{arXiv preprint arXiv:1506.02438}.

\bibitem[{Schulman et~al.(2017)Schulman, Wolski, Dhariwal, Radford, and
  Klimov}]{schulman2017proximal}
Schulman, J.; Wolski, F.; Dhariwal, P.; Radford, A.; and Klimov, O. 2017.
\newblock Proximal policy optimization algorithms.
\newblock \emph{arXiv preprint arXiv:1707.06347}.

\bibitem[{Shen et~al.(2018)Shen, Zhou, Long, Jiang, Wang, and
  Zhang}]{shen2018reinforced}
Shen, T.; Zhou, T.; Long, G.; Jiang, J.; Wang, S.; and Zhang, C. 2018.
\newblock Reinforced self-attention network: a hybrid of hard and soft
  attention for sequence modeling.
\newblock \emph{arXiv preprint arXiv:1801.10296}.

\bibitem[{Son et~al.(2019)Son, Kim, Kang, Hostallero, and Yi}]{son2019qtran}
Son, K.; Kim, D.; Kang, W.~J.; Hostallero, D.~E.; and Yi, Y. 2019.
\newblock Qtran: Learning to factorize with transformation for cooperative
  multi-agent reinforcement learning.
\newblock In \emph{International Conference on Machine Learning}, 5887--5896.
  PMLR.

\bibitem[{Sunehag et~al.(2017)Sunehag, Lever, Gruslys, Czarnecki, Zambaldi,
  Jaderberg, Lanctot, Sonnerat, Leibo, Tuyls et~al.}]{sunehag2017value}
Sunehag, P.; Lever, G.; Gruslys, A.; Czarnecki, W.~M.; Zambaldi, V.; Jaderberg,
  M.; Lanctot, M.; Sonnerat, N.; Leibo, J.~Z.; Tuyls, K.; et~al. 2017.
\newblock Value-decomposition networks for cooperative multi-agent learning.
\newblock \emph{arXiv preprint arXiv:1706.05296}.

\bibitem[{Sutton, Barto et~al.(1998)}]{sutton1998introduction}
Sutton, R.~S.; Barto, A.~G.; et~al. 1998.
\newblock \emph{Introduction to reinforcement learning}, volume 135.
\newblock MIT press Cambridge.

\bibitem[{Tampuu et~al.(2017)Tampuu, Matiisen, Kodelja, Kuzovkin, Korjus, Aru,
  Aru, and Vicente}]{tampuu2017multiagent}
Tampuu, A.; Matiisen, T.; Kodelja, D.; Kuzovkin, I.; Korjus, K.; Aru, J.; Aru,
  J.; and Vicente, R. 2017.
\newblock Multiagent cooperation and competition with deep reinforcement
  learning.
\newblock \emph{PloS one}, 12(4): e0172395.

\bibitem[{Usunier et~al.(2016)Usunier, Synnaeve, Lin, and
  Chintala}]{usunier2016episodic}
Usunier, N.; Synnaeve, G.; Lin, Z.; and Chintala, S. 2016.
\newblock Episodic exploration for deep deterministic policies: An application
  to starcraft micromanagement tasks.
\newblock \emph{arXiv preprint arXiv:1609.02993}.

\bibitem[{Vaswani et~al.(2017)Vaswani, Shazeer, Parmar, Uszkoreit, Jones,
  Gomez, Kaiser, and Polosukhin}]{vaswani2017attention}
Vaswani, A.; Shazeer, N.; Parmar, N.; Uszkoreit, J.; Jones, L.; Gomez, A.~N.;
  Kaiser, {\L}.; and Polosukhin, I. 2017.
\newblock Attention is all you need.
\newblock In \emph{Advances in neural information processing systems},
  5998--6008.

\bibitem[{Vinyals et~al.(2017)Vinyals, Ewalds, Bartunov, Georgiev, Vezhnevets,
  Yeo, Makhzani, K{\"u}ttler, Agapiou, Schrittwieser
  et~al.}]{vinyals2017starcraft}
Vinyals, O.; Ewalds, T.; Bartunov, S.; Georgiev, P.; Vezhnevets, A.~S.; Yeo,
  M.; Makhzani, A.; K{\"u}ttler, H.; Agapiou, J.; Schrittwieser, J.; et~al.
  2017.
\newblock Starcraft ii: A new challenge for reinforcement learning.
\newblock \emph{arXiv preprint arXiv:1708.04782}.

\bibitem[{Watkins(1989)}]{watkins1989learning}
Watkins, C. J. C.~H. 1989.
\newblock Learning from delayed rewards.

\bibitem[{Williams(1992)}]{williams1992simple}
Williams, R.~J. 1992.
\newblock Simple statistical gradient-following algorithms for connectionist
  reinforcement learning.
\newblock \emph{Machine learning}, 8(3): 229--256.

\bibitem[{Xu et~al.(2015)Xu, Ba, Kiros, Cho, Courville, Salakhudinov, Zemel,
  and Bengio}]{xu2015show}
Xu, K.; Ba, J.; Kiros, R.; Cho, K.; Courville, A.; Salakhudinov, R.; Zemel, R.;
  and Bengio, Y. 2015.
\newblock Show, attend and tell: Neural image caption generation with visual
  attention.
\newblock In \emph{International conference on machine learning}, 2048--2057.
  PMLR.

\bibitem[{Yang et~al.(2020)Yang, Hao, Chen, Tang, Chen, Hu, Fan, and
  Wei}]{yang2020q}
Yang, Y.; Hao, J.; Chen, G.; Tang, H.; Chen, Y.; Hu, Y.; Fan, C.; and Wei, Z.
  2020.
\newblock Q-value path decomposition for deep multiagent reinforcement
  learning.
\newblock In \emph{International Conference on Machine Learning}, 10706--10715.
  PMLR.

\bibitem[{Ye et~al.(2020)Ye, Liu, Sun, Shi, Zhao, Wu, Yu, Yang, Wu, Guo
  et~al.}]{ye2020mastering}
Ye, D.; Liu, Z.; Sun, M.; Shi, B.; Zhao, P.; Wu, H.; Yu, H.; Yang, S.; Wu, X.;
  Guo, Q.; et~al. 2020.
\newblock Mastering complex control in moba games with deep reinforcement
  learning.
\newblock In \emph{Proceedings of the AAAI Conference on Artificial
  Intelligence}, volume~34, 6672--6679.

\bibitem[{Zha et~al.(2021)Zha, Xie, Ma, Zhang, Lian, Hu, and
  Liu}]{zha2021douzero}
Zha, D.; Xie, J.; Ma, W.; Zhang, S.; Lian, X.; Hu, X.; and Liu, J. 2021.
\newblock DouZero: Mastering DouDizhu with Self-Play Deep Reinforcement
  Learning.
\newblock \emph{arXiv preprint arXiv:2106.06135}.

\bibitem[{Zheng et~al.(2018)Zheng, Yang, Cai, Zhou, Zhang, Wang, and
  Yu}]{zheng2018magent}
Zheng, L.; Yang, J.; Cai, H.; Zhou, M.; Zhang, W.; Wang, J.; and Yu, Y. 2018.
\newblock Magent: A many-agent reinforcement learning platform for artificial
  collective intelligence.
\newblock In \emph{Proceedings of the AAAI Conference on Artificial
  Intelligence}, volume~32.

\bibitem[{Zhou et~al.(2021)Zhou, Zhang, Peng, Zhang, Li, Xiong, and
  Zhang}]{zhou2021informer}
Zhou, H.; Zhang, S.; Peng, J.; Zhang, S.; Li, J.; Xiong, H.; and Zhang, W.
  2021.
\newblock Informer: Beyond efficient transformer for long sequence time-series
  forecasting.
\newblock In \emph{Proceedings of AAAI}.

\end{thebibliography}

\end{document}